%% file: example_paper.tex
\documentclass{article}

\usepackage{microtype}
\usepackage{graphicx}
\usepackage{subcaption}
\usepackage{booktabs}
\usepackage{booktabs}  
\usepackage{multirow} 
\usepackage{graphicx} 
\usepackage{hyperref}

\usepackage{booktabs}  
\usepackage{amssymb}
\usepackage[table]{xcolor} 

\usepackage{tabularx}
\usepackage[preprint]{icml2026}

\usepackage{amsmath}
\usepackage{amssymb}
\usepackage{mathtools}
\usepackage{amsthm}
\usepackage{verbatim}

\usepackage[capitalize,noabbrev]{cleveref}

\theoremstyle{plain}

\theoremstyle{definition}

\theoremstyle{remark}

\icmltitlerunning{RelayCaching}

\begin{document}

\twocolumn[
  \icmltitle{RelayCaching: Accelerating LLM Collaboration via Decoding KV Cache Reuse}

  \icmlsetsymbol{equal}{*}

  \begin{icmlauthorlist}
    \icmlauthor{Yingsheng Geng}{bupt}
    \icmlauthor{Yuchong Gao}{thu}
    \icmlauthor{Weihong Wu}{uestc}
    \icmlauthor{Guyue Liu}{pku}
    \icmlauthor{Jiang Liu}{bupt}
  \end{icmlauthorlist}

  \icmlaffiliation{bupt}{Beijing University of Posts and Telecommunications, Beijing, China}
  \icmlaffiliation{thu}{Tsinghua University, Beijing, China}
  \icmlaffiliation{uestc}{University of Electronic Science and Technology of China, Chengdu, China}
  \icmlaffiliation{pku}{Peking University, Beijing, China}
    
  \icmlcorrespondingauthor{Weihong Wu}{wuweihong@uestc.edu.cn}
  \icmlcorrespondingauthor{Guyue Liu}{guyue@pku.edu.cn}
  
  \icmlkeywords{Large Language Models, Multi-agent Systems, KV Cache, Decoding Acceleration, Cache Reuse}

  \vskip 0.3in
]

\printAffiliationsAndNotice{}

\begin{abstract}
The increasing complexity of AI tasks has shifted the paradigm from monolithic models toward multi-agent large language model (LLM) systems.
However, these collaborative architectures introduce a critical bottleneck: redundant prefill computation for shared content generated by previous agents, which significantly increases KV cache memory usage and time-to-first-token (TTFT). 
While various KV cache methods have been proposed to mitigate prefill redundancy, they either fail to maintain accuracy on agent-generated outputs or exhibit low reuse rates due to rigid constraints.
We present RelayCaching, a training-free inference method that directly reuses decoding phase KV caches from previous agents in subsequent prefill phases. Our key insight is that KV caches for identical content are highly consistent across phases, while prefix-induced deviations are sparse and localized within a limited range of layers and token positions. By selectively recomputing KV caches at these positions, RelayCaching preserves model accuracy with minimal overhead, yielding a superior accuracy–efficiency trade-off over existing methods.
Experiments on diverse collaborative LLM tasks spanning mathematical reasoning, general knowledge, and code generation demonstrate that RelayCaching achieves over $80$\% KV cache reuse, reduces TTFT by up to $4.7\times$ compared to the standard pipeline, all with negligible accuracy degradation.
\end{abstract}

\input{contents/1-intro}
\input{contents/2-background}
\input{contents/3-observation}
\input{contents/4-relaycaching}

\input{contents/5-evaluation}
\input{contents/6-conclusion}

\section*{Impact Statement}
This paper presents work whose goal is to advance the field of Machine Learning. There are many potential societal consequences of our work, none which we feel must be specifically highlighted here.

\bibliography{bib/reference}
\bibliographystyle{icml2026}

%%%%%%%%%%%%%%%%%%%%%%%%%%%%%%%%%%%%%%%%%%%%%%%%%%%%%%%%%%%%%%%%%%%%%%%%%%%%%%%
%%%%%%%%%%%%%%%%%%%%%%%%%%%%%%%%%%%%%%%%%%%%%%%%%%%%%%%%%%%%%%%%%%%%%%%%%%%%%%%
% APPENDIX
%%%%%%%%%%%%%%%%%%%%%%%%%%%%%%%%%%%%%%%%%%%%%%%%%%%%%%%%%%%%%%%%%%%%%%%%%%%%%%%
%%%%%%%%%%%%%%%%%%%%%%%%%%%%%%%%%%%%%%%%%%%%%%%%%%%%%%%%%%%%%%%%%%%%%%%%%%%%%%%
\newpage
\appendix
\onecolumn
\section{Further Observations}
\label{app:further_observations}

In Section~\ref{sec:observation}, we presented the systematic deviation patterns of decoding-to-prefill KV reuse using Mistral-7B-Instruct-v0.3. To verify the universality of these phenomena across different model architectures, sizes, and capabilities, we extend our analysis to five additional state-of-the-art models: 
Llama-3.1-8B-Instruct, 
Qwen2.5-Coder-7B-Instruct, 
DeepSeek-R1-Distill-Qwen-32B, 
Qwen3-30B-A3B, and the lightweight 
Qwen3-0.6B.

\begin{figure*}[p] 
    \centering
    \setlength{\tabcolsep}{1pt} % Reduce gap between columns
    \begin{tabular}{cc}
        \textbf{Layer-wise Similarity (U-Shape)} & \textbf{Relative Recovery Curves} \\
        
        % --- Row 1: Llama-3.1-8B ---
        \includegraphics[width=0.48\linewidth]{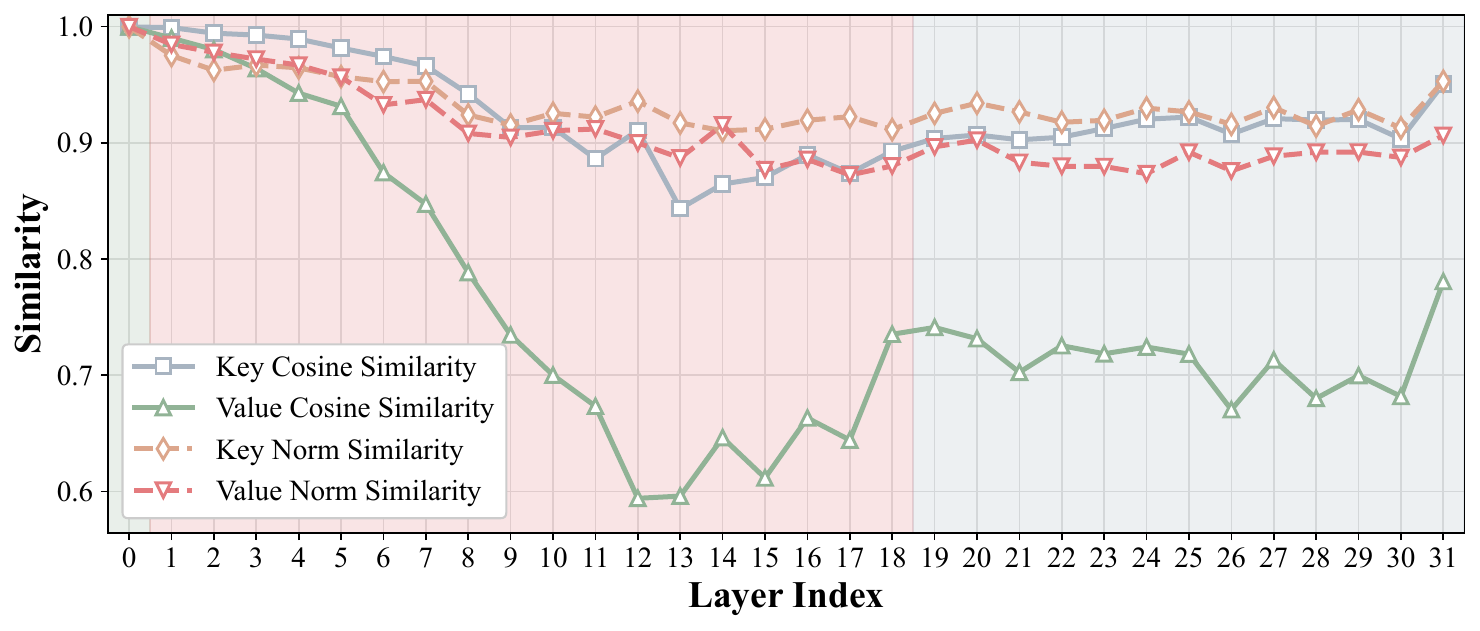} &
        \includegraphics[width=0.48\linewidth]{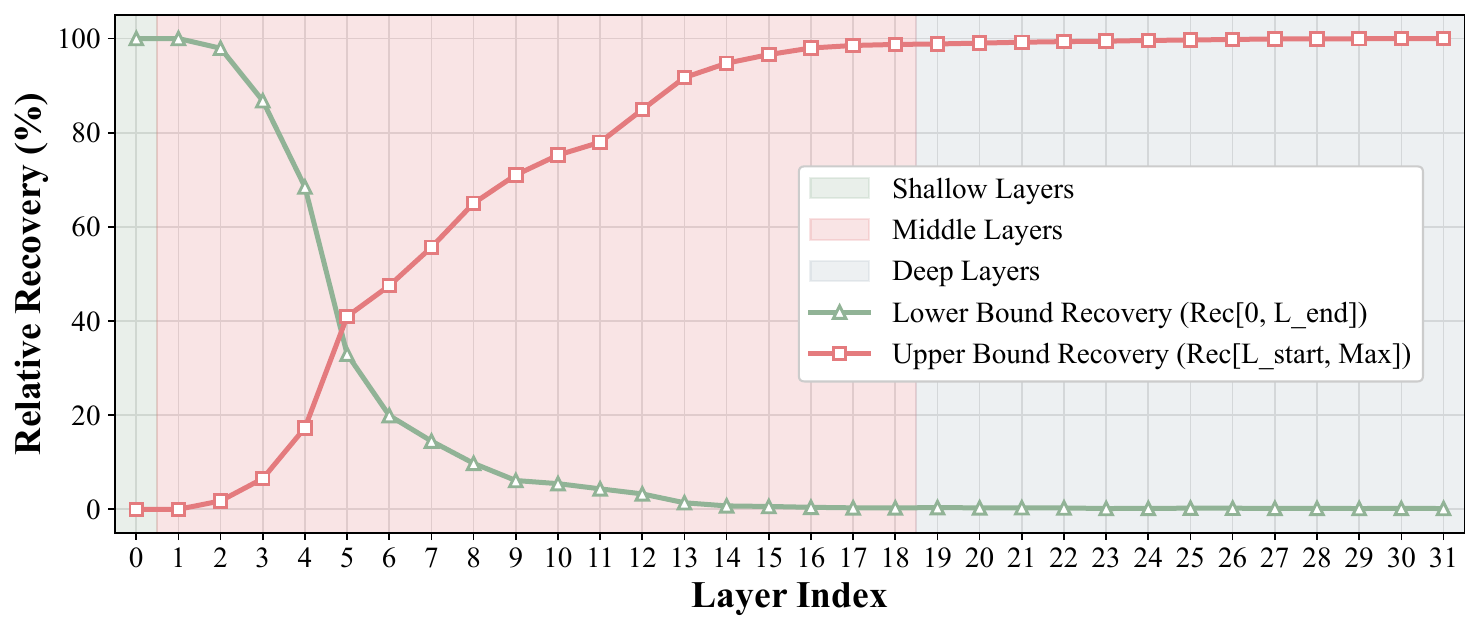} \\
        \multicolumn{2}{c}{\small (a) Llama-3.1-8B-Instruct} \\[\smallskipamount]
        
        % --- Row 2: Qwen2.5-Coder-7B ---
        \includegraphics[width=0.48\linewidth]{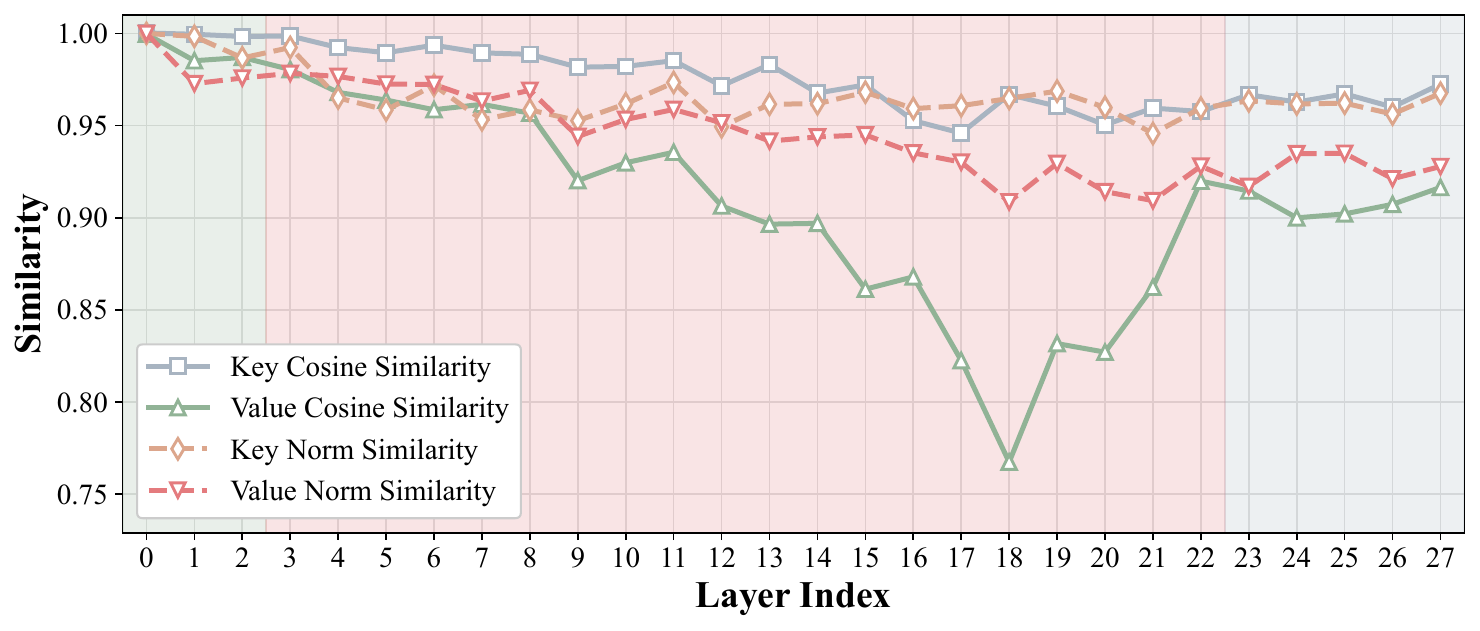} &
        \includegraphics[width=0.48\linewidth]{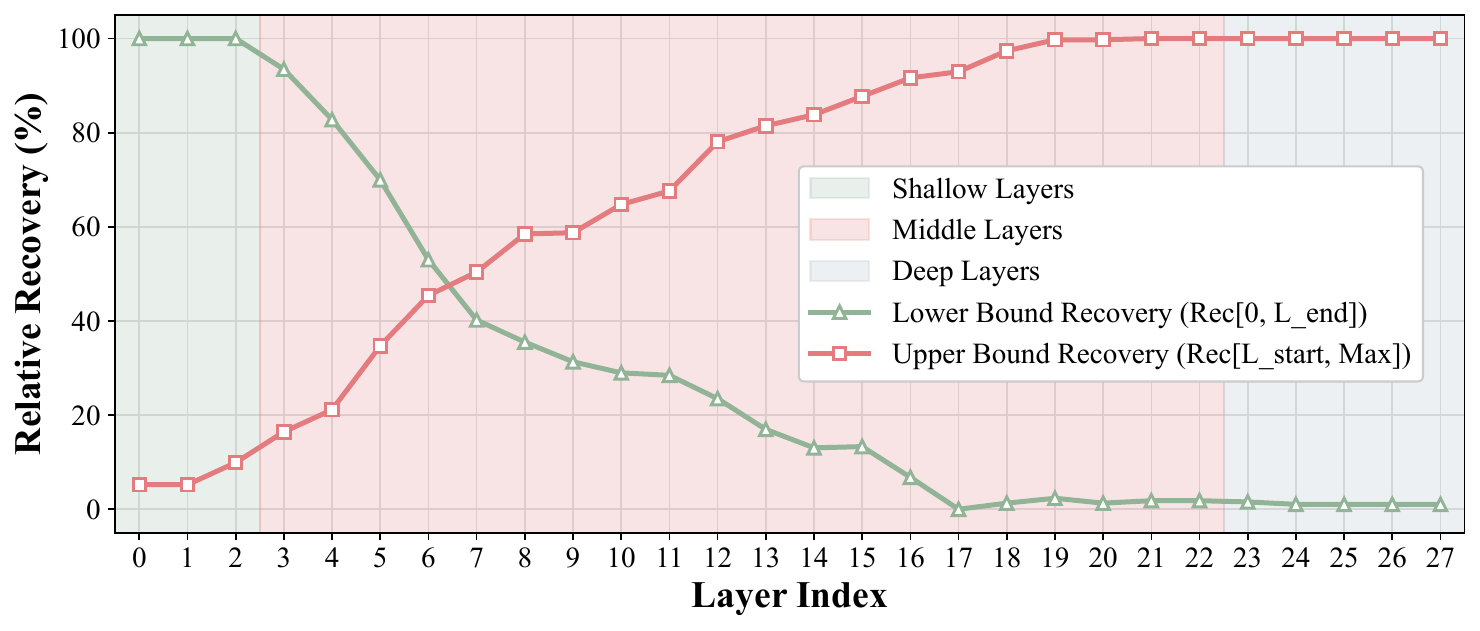} \\
        \multicolumn{2}{c}{\small (b) Qwen2.5-Coder-7B-Instruct} \\[\smallskipamount]
        
        % --- Row 3: DeepSeek-R1 ---
        \includegraphics[width=0.48\linewidth]{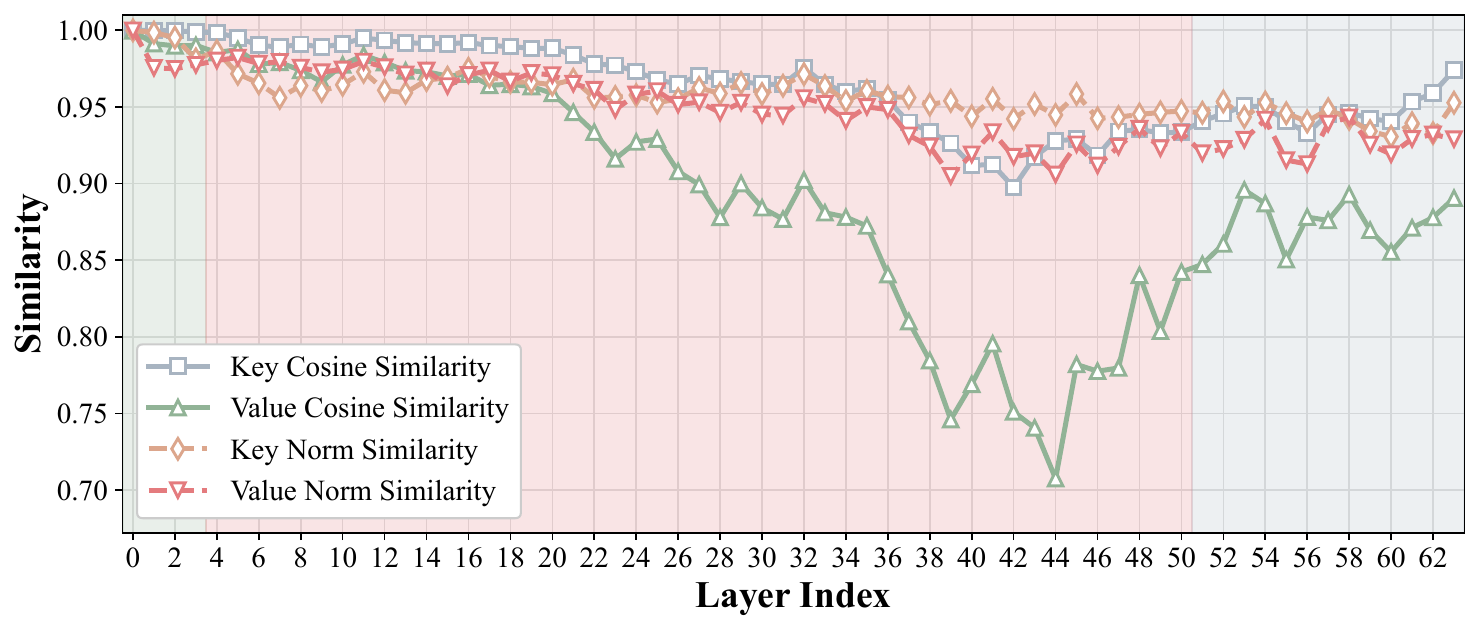} &
        \includegraphics[width=0.48\linewidth]{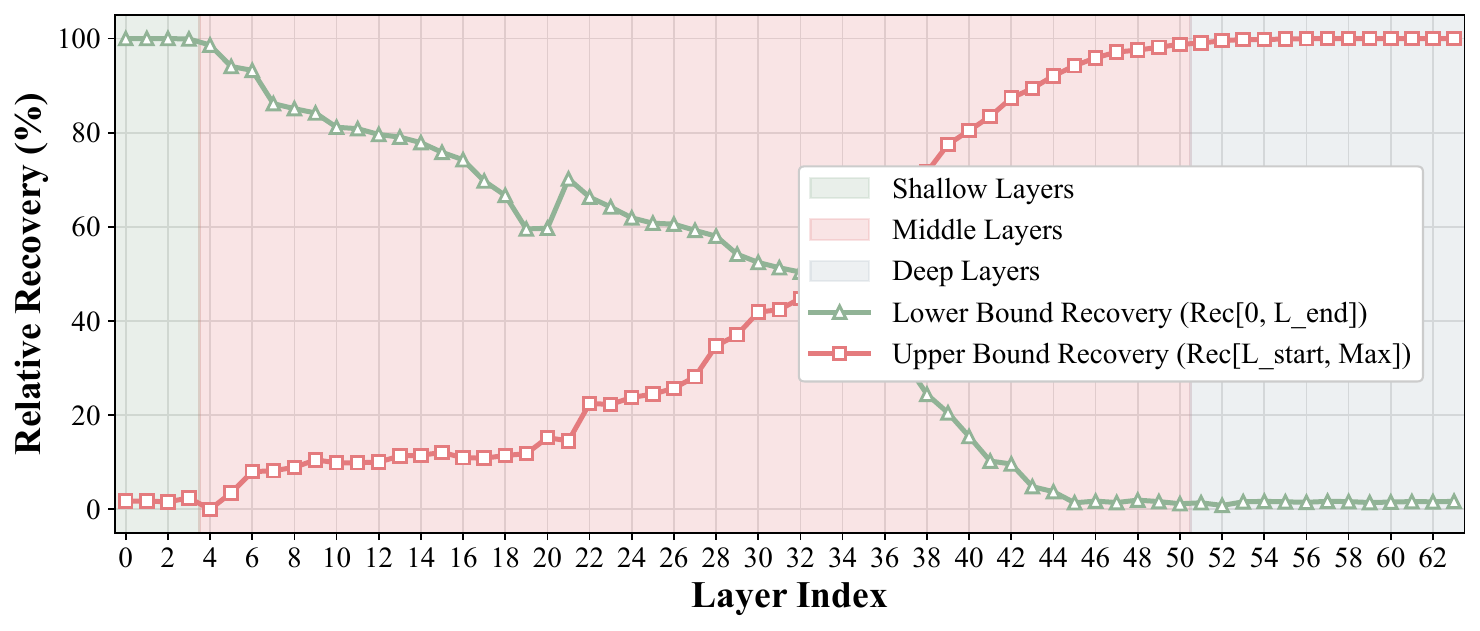} \\
        \multicolumn{2}{c}{\small (c) DeepSeek-R1-Distill-Qwen-32B} \\[\smallskipamount]

        % --- Row 4: Qwen3-30B ---
        \includegraphics[width=0.48\linewidth]{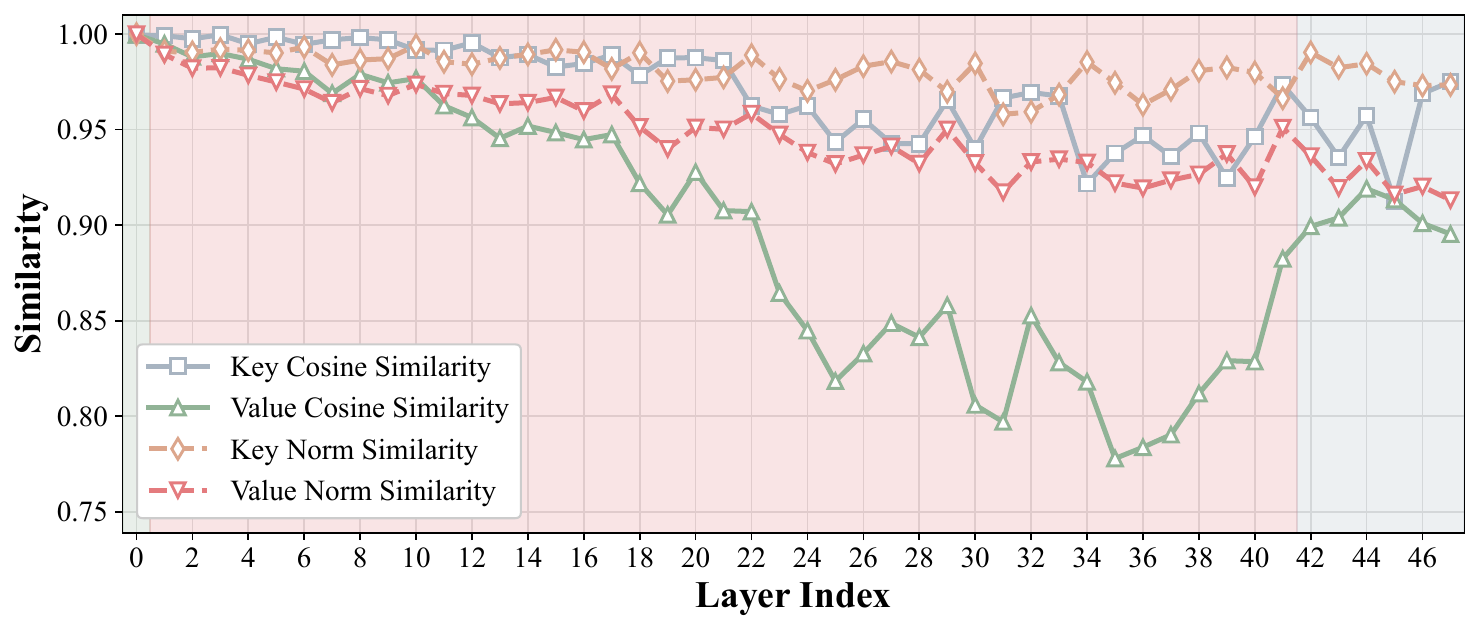} &
        \includegraphics[width=0.48\linewidth]{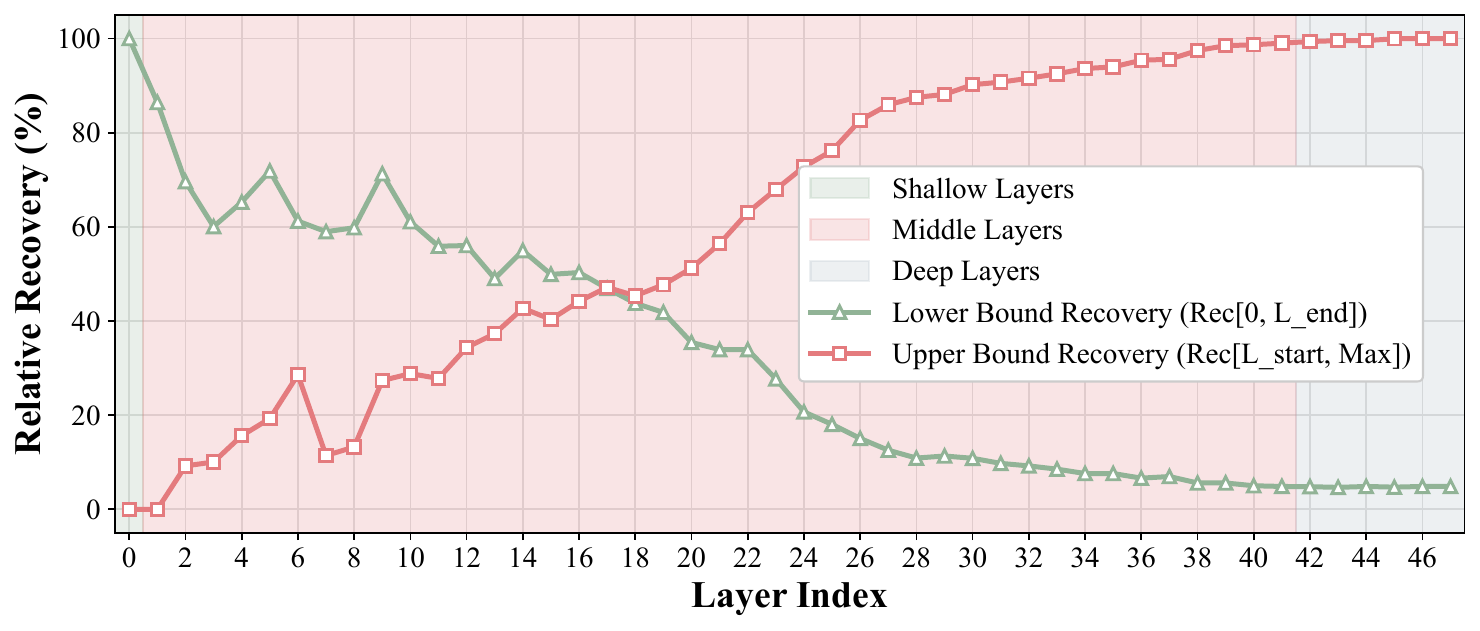} \\
        \multicolumn{2}{c}{\small (d) Qwen3-30B-A3B} \\[\smallskipamount]

        % --- Row 5: Qwen3-0.6B ---
        \includegraphics[width=0.48\linewidth]{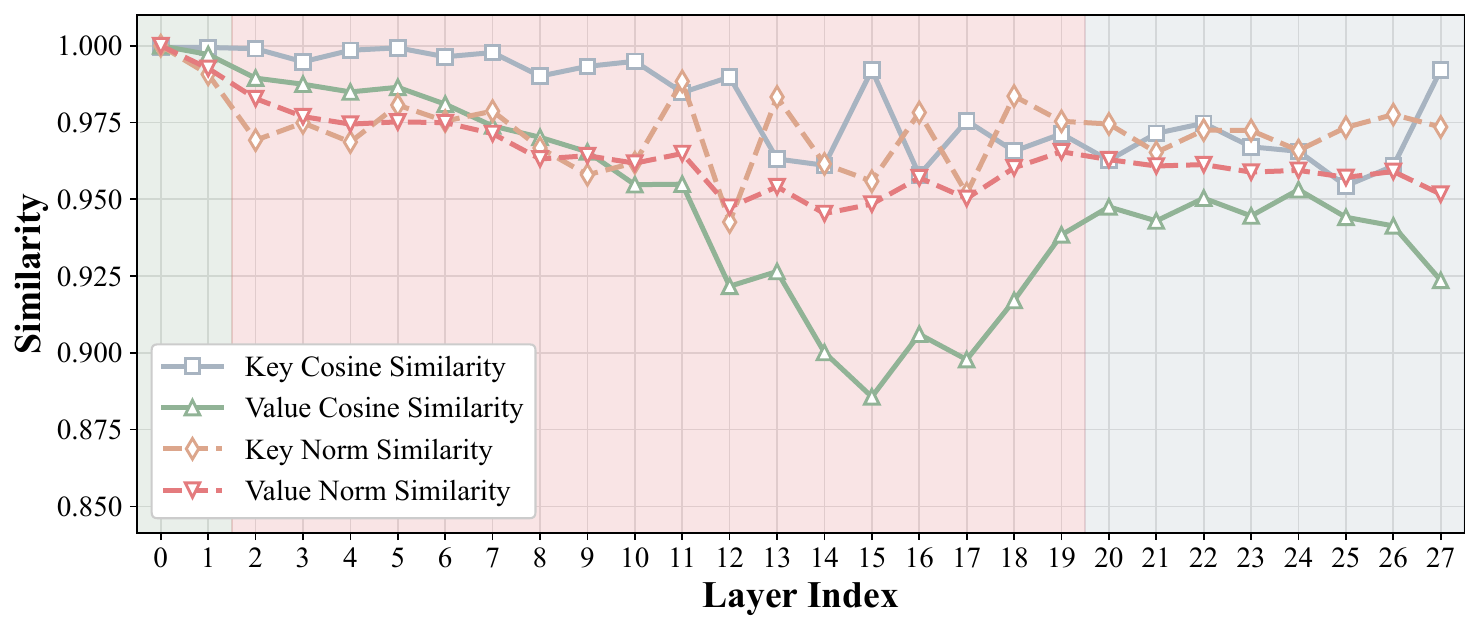} &
        \includegraphics[width=0.48\linewidth]{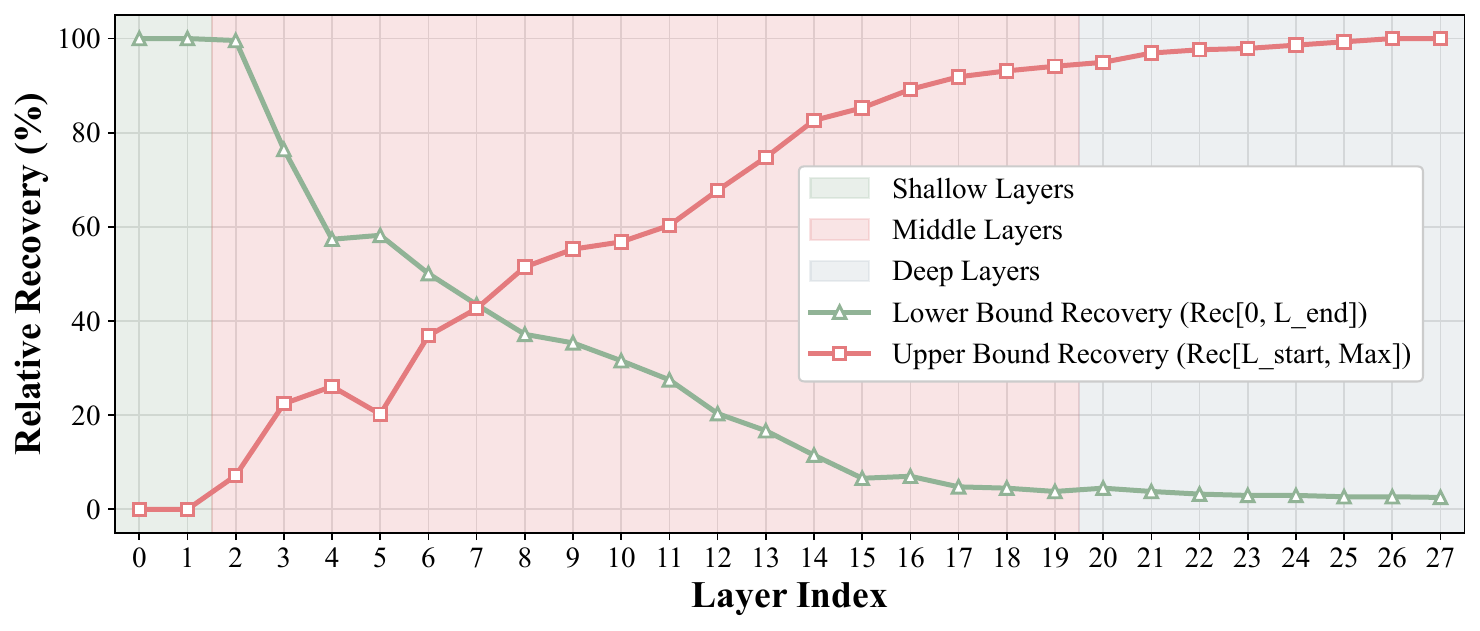} \\
        \multicolumn{2}{c}{\small (e) Qwen3-0.6B} \\
        
    \end{tabular}
    \caption{\textbf{Universality of layer-wise structured deviations.} 
    \textbf{Left Column:} Layer-wise cosine similarity and norm similarity profiles. The distinct U-shaped drop in value cosine similarity (red lines) is consistent across diverse models.
    \textbf{Right Column:} Relative recovery curves. In all cases, recomputing the middle layers (where the similarity drops) yields the most efficient recovery of downstream token fidelity.}
    \label{fig:app_layer_patterns}
\end{figure*}

\textbf{Universality of Layer-Wise Patterns.} 
Figure~\ref{fig:app_layer_patterns} illustrates the layer-wise cosine similarity profiles and relative recovery curves for these models.
Consistent with our main observations, all models exhibit a characteristic U-shaped value cosine similarity profile. The "middle layers"---where reasoning and context integration primarily occur---consistently  show the highest sensitivity to prefix shifts (lowest similarity), while shallow layers maintain relatively high fidelity and deep layers only partially recover.

Furthermore, the recovery curves confirm that this deviation is structured: rectifying the specific range of middle layers identified by the U-curve yields the steepest similarity recovery across all models. For instance, even in the larger DeepSeek-R1-Distill-Qwen-32B or the architectural variant Qwen3-30B-A3B, the strategy of prioritizing middle-layer rectification remains the most effective.

\begin{figure*}[p]
    \centering
    \setlength{\tabcolsep}{1pt}
    \begin{tabular}{cc}
        \textbf{Token-wise Similarity (Sparsity)} & \textbf{Rank Correlation (Persistence)} \\
        
        % --- Row 1: Llama-3.1-8B ---
        \includegraphics[width=0.48\linewidth]{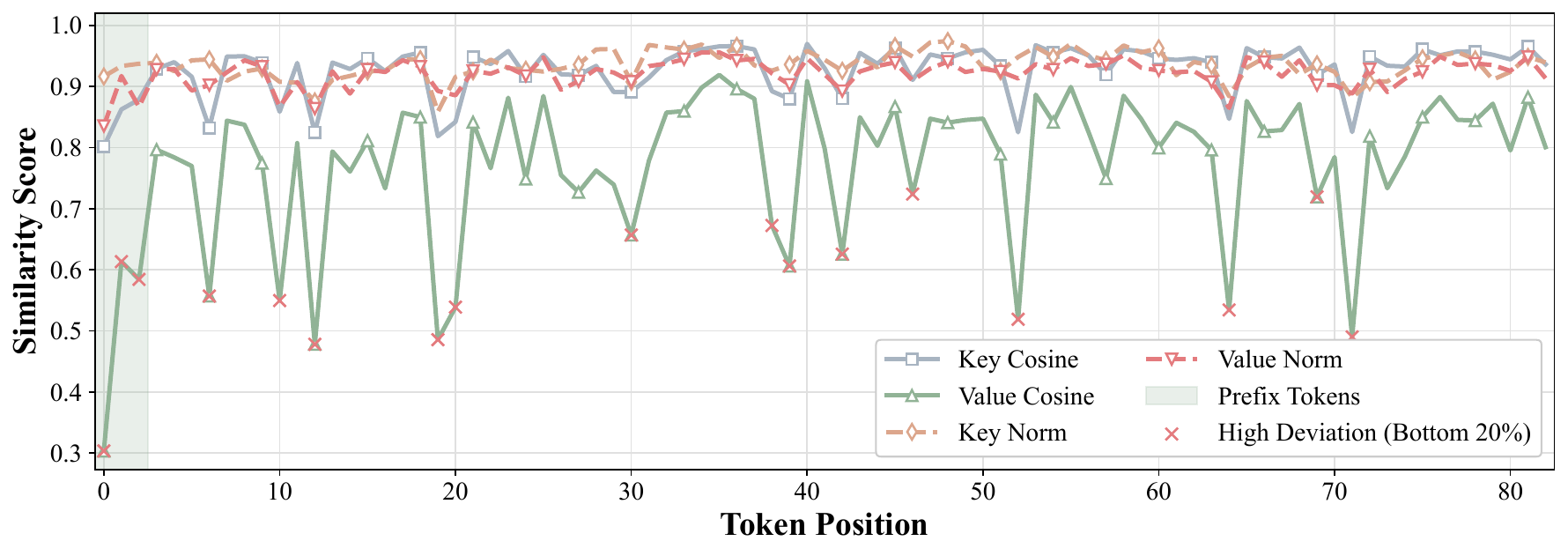} &
        \includegraphics[width=0.48\linewidth]{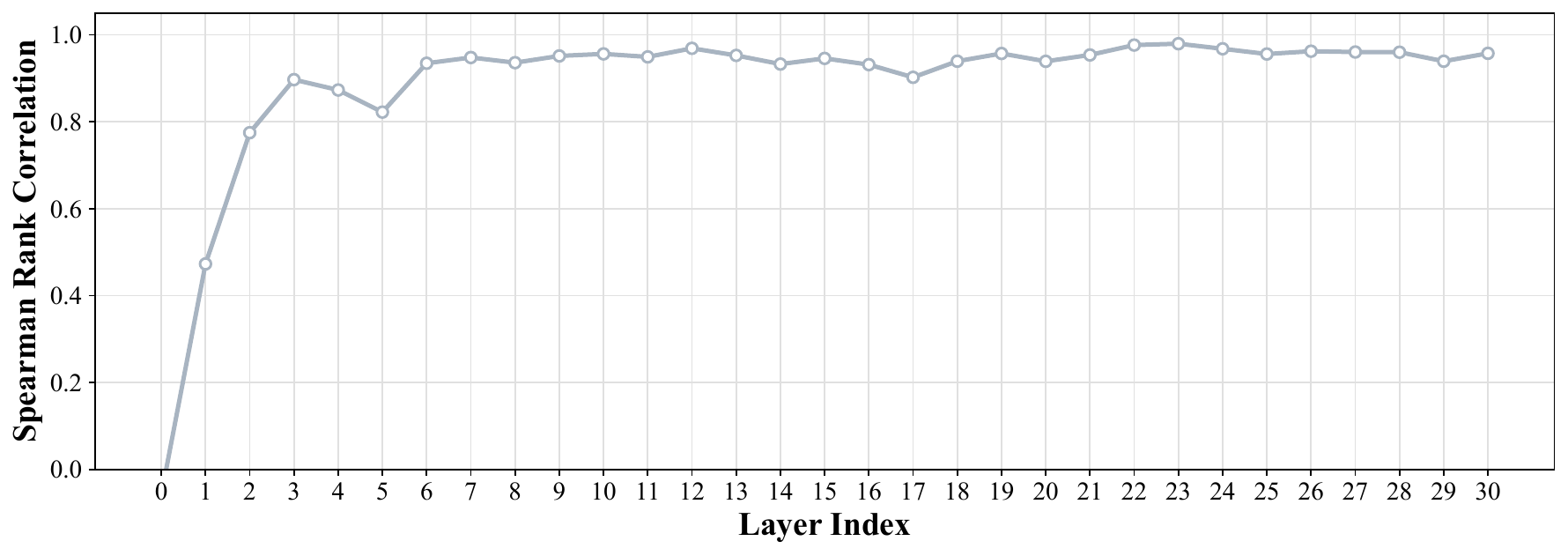} \\
        \multicolumn{2}{c}{\small (a) Llama-3.1-8B-Instruct} \\[\smallskipamount]
        
        % --- Row 2: Qwen2.5-Coder-7B ---
        \includegraphics[width=0.48\linewidth]{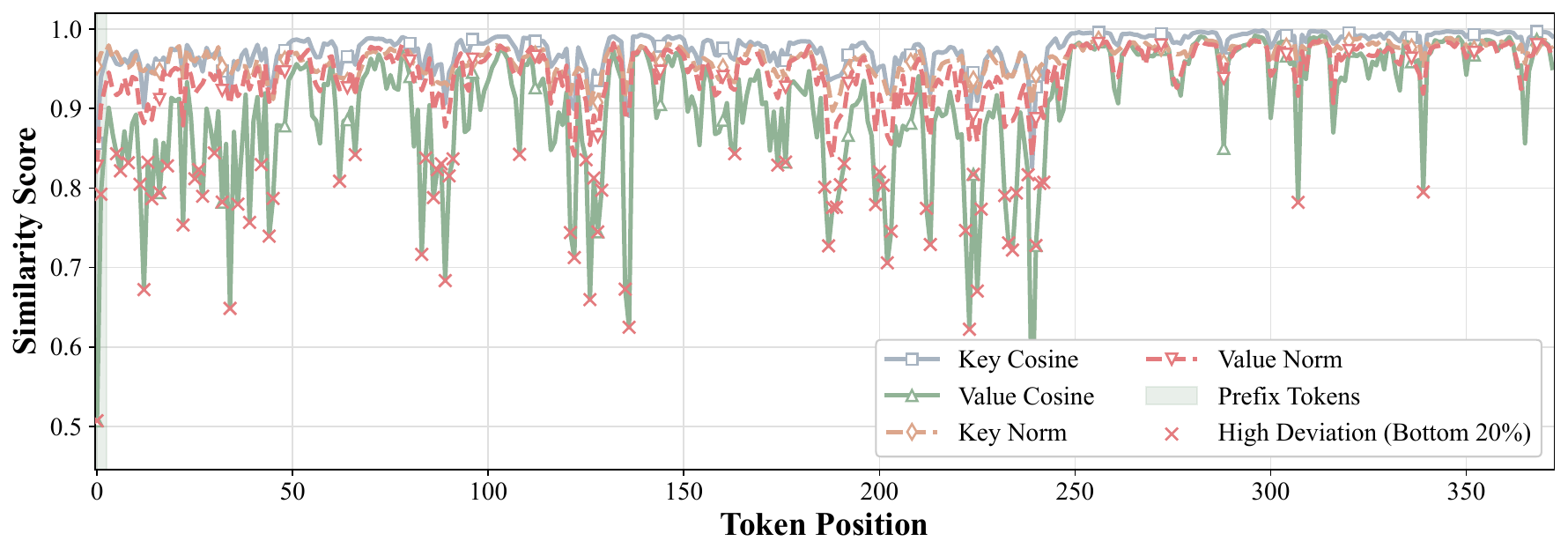} &
        \includegraphics[width=0.48\linewidth]{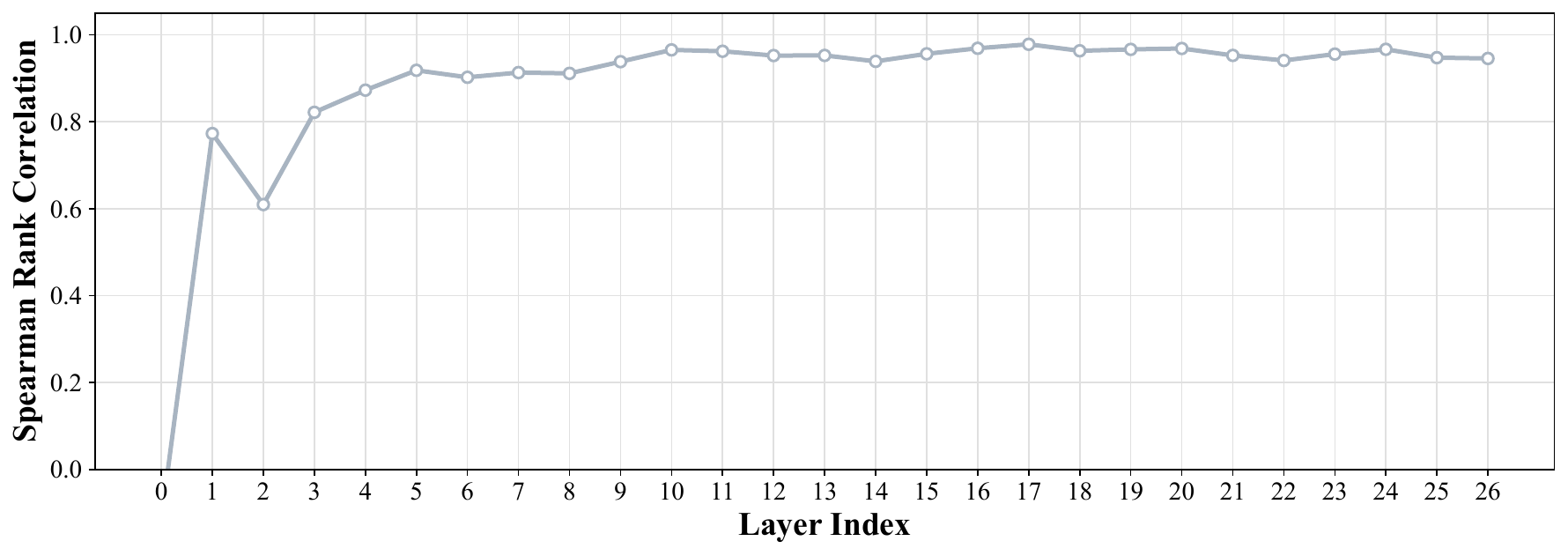} \\
        \multicolumn{2}{c}{\small (b) Qwen2.5-Coder-7B-Instruct} \\[\smallskipamount]
        
        % --- Row 3: DeepSeek-R1 ---
        \includegraphics[width=0.48\linewidth]{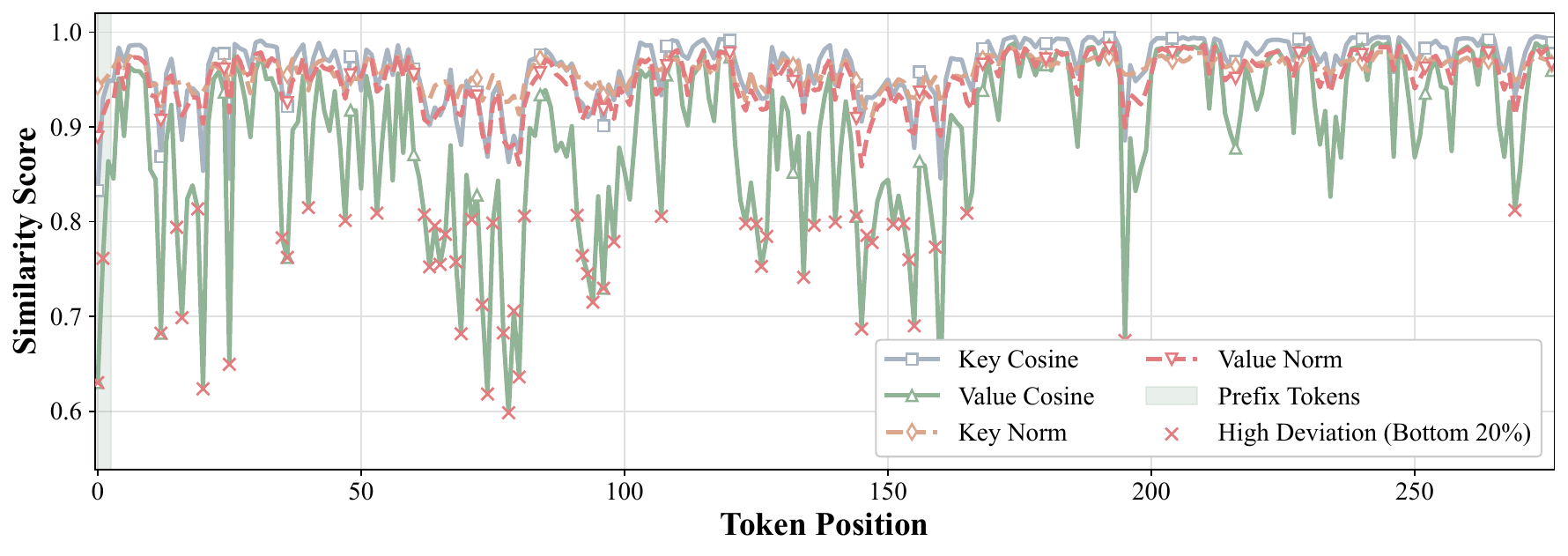} &
        \includegraphics[width=0.48\linewidth]{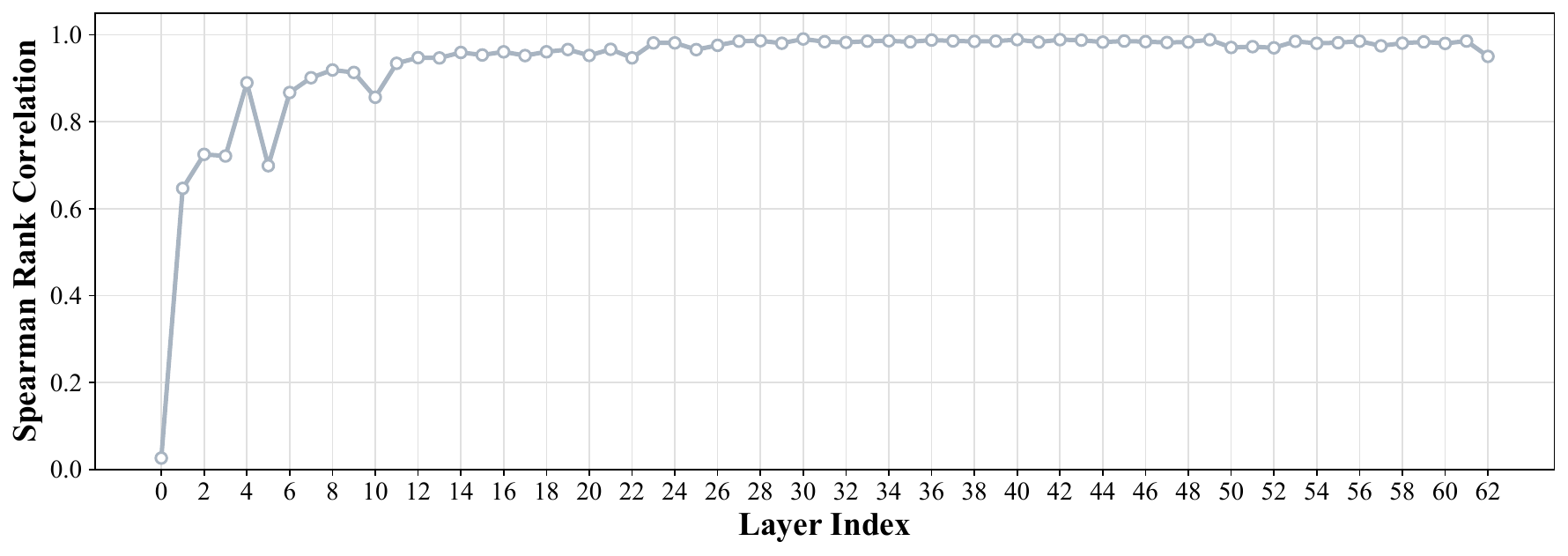} \\
        \multicolumn{2}{c}{\small (c) DeepSeek-R1-Distill-Qwen-32B} \\[\smallskipamount]

        % --- Row 4: Qwen3-30B ---
        \includegraphics[width=0.48\linewidth]{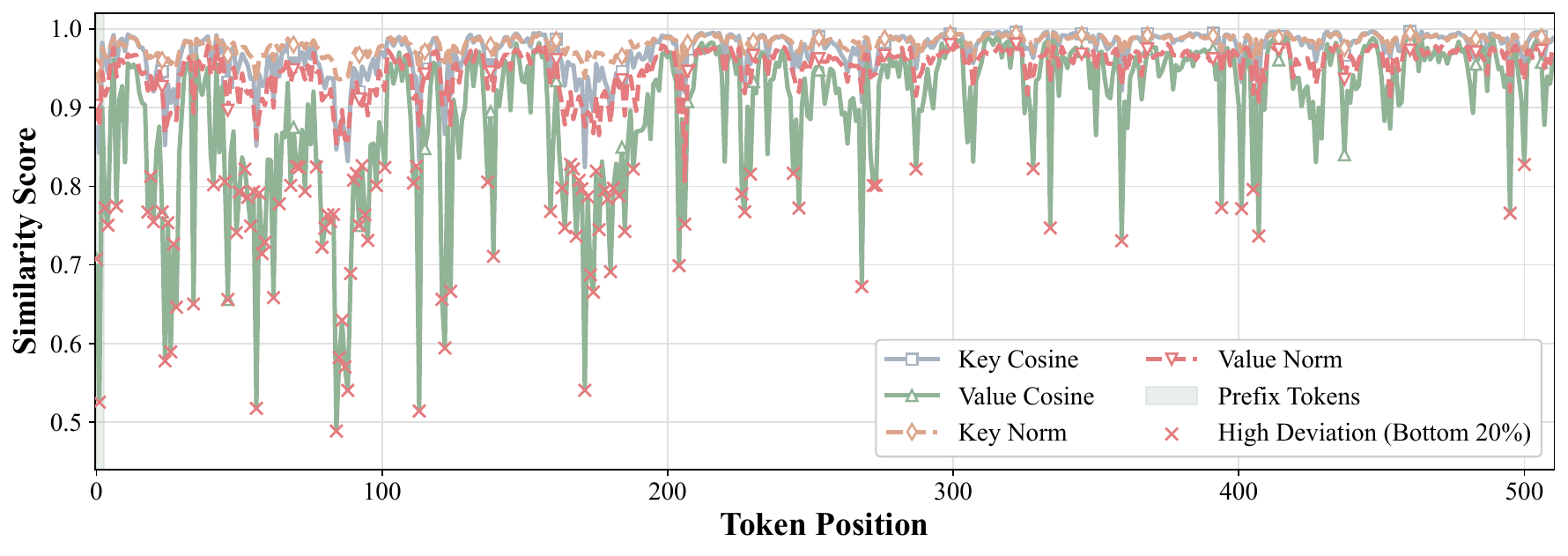} &
        \includegraphics[width=0.48\linewidth]{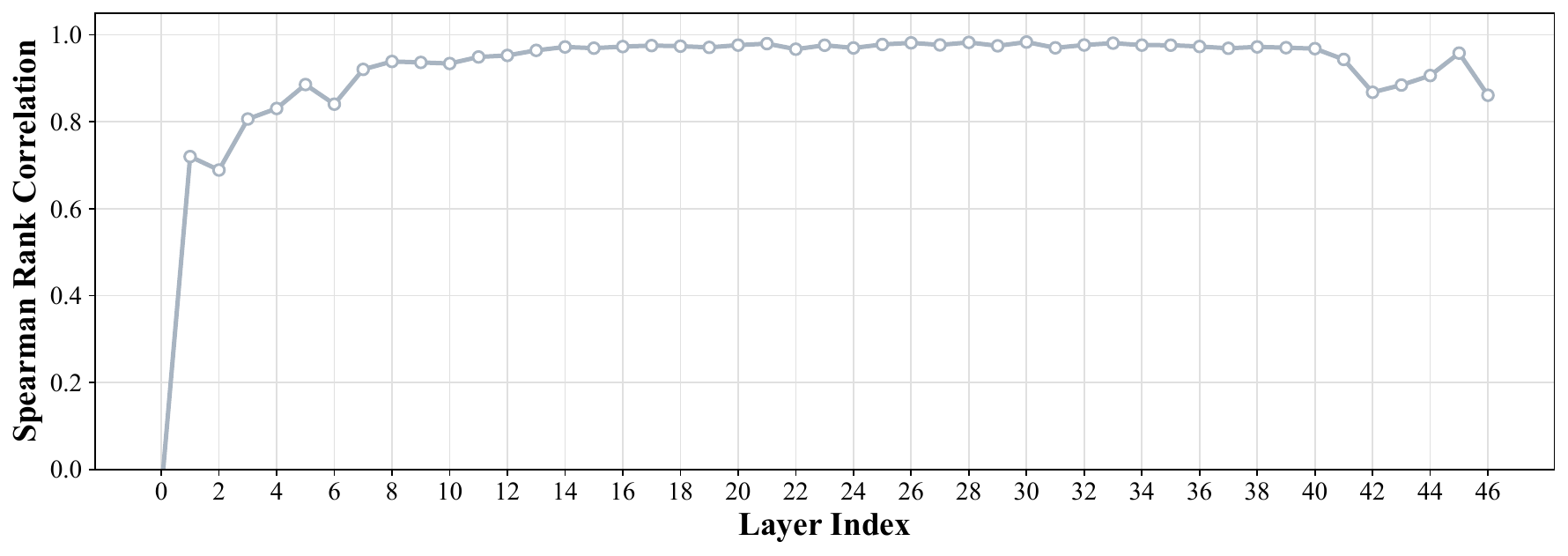} \\
        \multicolumn{2}{c}{\small (d) Qwen3-30B-A3B} \\[\smallskipamount]

        % --- Row 5: Qwen3-0.6B ---
        \includegraphics[width=0.48\linewidth]{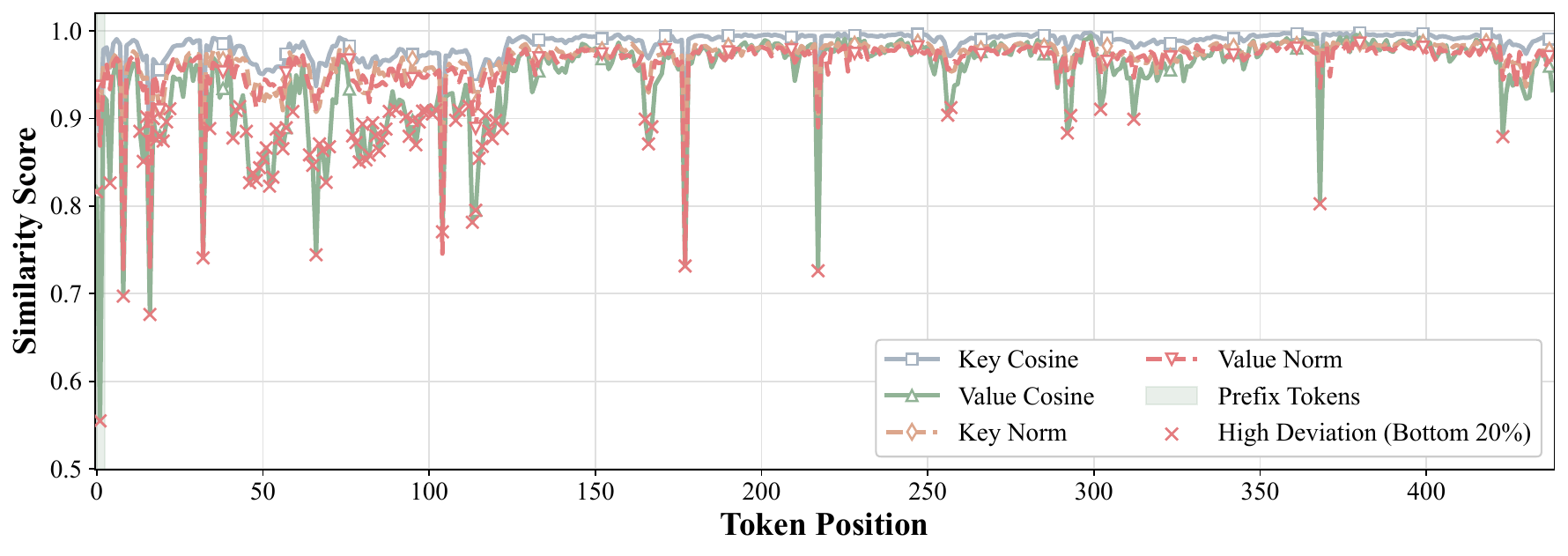} &
        \includegraphics[width=0.48\linewidth]{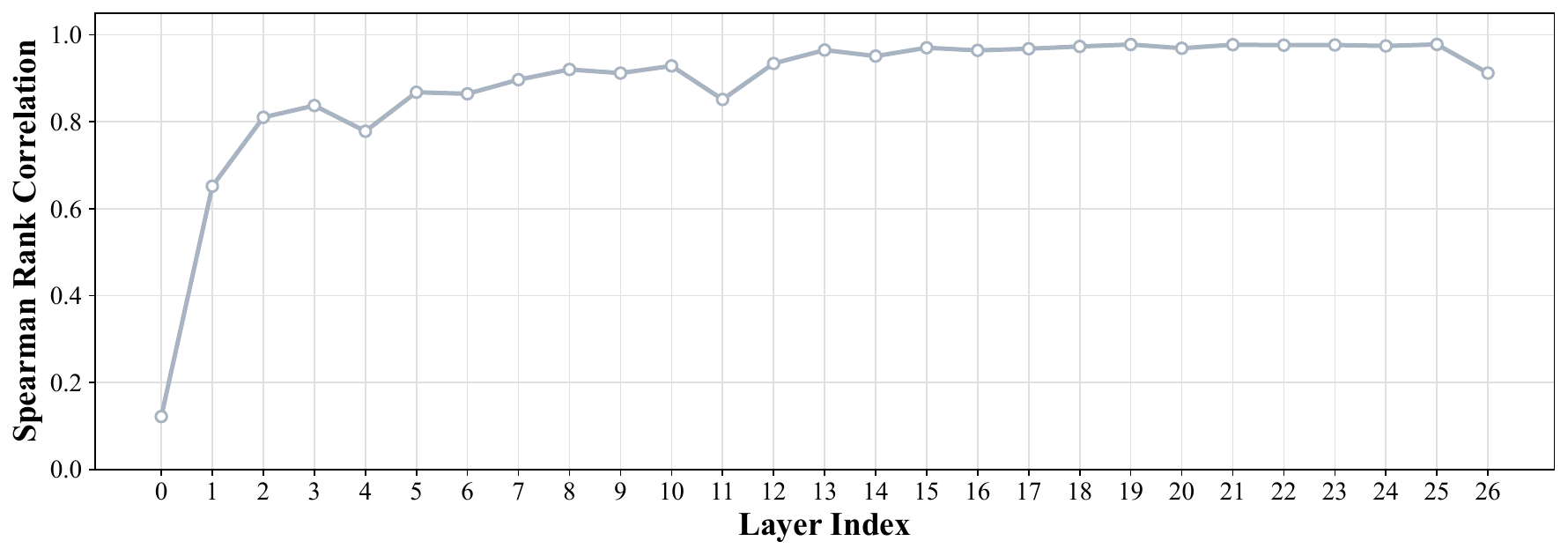} \\
        \multicolumn{2}{c}{\small (e) Qwen3-0.6B} \\
        
    \end{tabular}
    \caption{\textbf{Universality of token-wise structured deviations.}
    \textbf{Left Column:} Token-wise similarity profiles averaged over middle layers. The deviations are consistently sparse, with most tokens retaining high fidelity.
    \textbf{Right Column:} Spearman rank correlation of value cosine deviation between adjacent layers. The high correlation confirms that deviation patterns are persistent and predictable across depths.}
    \label{fig:app_token_patterns}
\end{figure*}

\textbf{Universality of Token-Wise Patterns.}
Figure~\ref{fig:app_token_patterns} depicts the token-wise deviation distribution and inter-layer rank correlations.
Across all evaluated models, we observe structured sparsity: only a small subset of tokens exhibits significant value deviation, while the majority remain highly aligned with full-prefill states.
Crucially, the rank correlation analysis reveals that deviations are not random noise. The high Spearman rank correlation between adjacent layers (rapidly stabilizing $>0.9$) indicates that once a token is corrupted by prefix shifts in early layers, it tends to remain a high-deviation outlier in subsequent layers. This persistence is observed universally, from the 0.6B model to the 32B model, validating the robustness of our selection-based rectification strategy.

% =========================================================
% Figure A: Layer-wise Patterns (5 Models x 2 Plots)
% =========================================================

% =========================================================
% Figure B: Token-wise Patterns (5 Models x 2 Plots)
% =========================================================
% \section{Implementation Details}
\section{Experiment Details}
\label{app:experiment_details}

In this section, we provide detailed configurations for our observation experiments, baseline implementations, and the main evaluation setup to ensure reproducibility.

\subsection{Setup for Observation Experiment}
\label{app:obs_setup}

To investigate the alignment between decoding and prefilling KV caches (Section~\ref{sec:observation}), we constructed a controlled \textit{Summarize-then-Answer} pipeline using the 2WikiMQA dataset~\cite{ho2020constructing}. 

\textbf{Data Construction.} We utilized the validation set processed into a JSON format. The dataset consists of 200 samples. For each instance, we formulated a two-stage process:
\begin{enumerate}
    \item \textbf{Stage 1 (Decoding):} The model generates a summary for a given Wikipedia passage. The maximum number of generated tokens is set to 512. The KV caches generated during this decoding phase are stored as the 
    \textit{Decoding KV}.
    \item \textbf{Stage 2 (Prefilling):} The generated summary is concatenated with a new prefix (derived from the question and task). We perform a full prefill computation on this sequence to obtain the ground-truth \textit{Full-Prefill} KV cache. 
\end{enumerate}

\textbf{Configuration.} We utilized Mistral-7B-Instruct-v0.3 as the primary probe model for the observation analysis presented in the main text (Section~\ref{sec:observation}). Additional models (e.g., Qwen, Llama, DeepSeek) analyzed in Appendix~\ref{app:further_observations} follow the same pipeline. All experiments were conducted in bfloat16 precision with greedy decoding (temperature $= 0$).

\subsection{RelayCaching Configuration}
\label{app:relay_config}

\textbf{Layer Selection.} The recompute layer range $[L_{\text{start}}, L_{\text{end}}]$ is determined offline using the profiling algorithm described in Section~\ref{sec:method}. The hyperparameters for this selection process are robust across models. We list the default values used in our experiments in Table~\ref{tab:layer_params}.

\begin{table}[h]
\centering
\caption{Hyperparameters for Critical Layer Selection. These values are used to identify the deviation-prone middle layers based on the U-shaped similarity profile.}
\label{tab:layer_params}
\begin{tabular}{lcc}
\toprule
Symbol & Description & Default \\
\midrule
$\tau_{\text{low}}$ & Lower similarity threshold & 0.99 \\
$T$ & Right-tail layers & 5 \\
$\lambda$ & Stable multiplier & 2.0 \\
$N_{\min}$ & Min rise samples & 3 \\
$C$ & Consecutive stables & 2 \\
\bottomrule
\end{tabular}
\end{table}

\textbf{Token Selection.} For the runtime token rectification, we adopt the following default settings based on our sensitivity studies (Figure~\ref{fig:hyperparameters}). These values were selected to maximize the reuse rate while maintaining generation quality across both reasoning and coding tasks:

\begin{itemize}
    \item \textbf{Difference Threshold ($\tau_{\text{diff}}$):} Set to 1.5. Tokens with a normalized KV difference score above this threshold times the mean deviation score are identified as high-deviation outliers and selected for rectification.
    \item \textbf{Downstream Threshold ($\tau_{\text{down}}$):} Set to 1.45. This threshold filters tokens based on their cumulative attention influence, ensuring that semantically pivotal tokens are rectified.
    \item \textbf{Suffix Length ($L_{\text{suf}}$):} Set to 10. We empirically found that rectifying the last 10 tokens of the relay segment is sufficient to correct local attention patterns and mitigate boundary artifacts.
\end{itemize}

\textbf{Model-Specific Layer Configurations.}
The specific critical layer boundaries identified by our profiling algorithm vary across model architectures. Table~\ref{tab:model_specific_layers} details the configurations used in our experiments. Specifically, $L_{\text{start}}$ denotes the reuse hidden layer where we begin injecting decoded states, and $L_{\text{det}}$ marks the detection layer used for computing difference metrics. $L_{\text{end}}$ marks the end of the critical range.

\begin{table}[h]
\centering
\caption{Model-specific layer configurations derived from offline profiling.}
\label{tab:model_specific_layers}
\begin{tabular}{lccc}
\toprule
\textbf{Model} & $L_{\text{start}}$ & $L_{\text{det}}$ & $L_{\text{end}}$ \\
\midrule
Llama-3.1-8B-Instruct           & 1  & 3  & 18 \\
Qwen2.5-Coder-7B-Instruct       & 3  & 4  & 22 \\
Qwen3-0.6B                      & 2  & 3  & 19 \\
\bottomrule
\end{tabular}
\end{table}

\subsection{Baseline Descriptions and Implementation}
\label{app:baselines}

We compare RelayCaching against performance bounds and state-of-the-art KV reuse methods. All baselines are implemented using the Hugging Face Transformers library.

\textbf{Performance Bounds:}
\begin{itemize}
    \item \textbf{Full-prefill (FULL):} The standard attention mechanism that performs complete prefilling for all input tokens, establishing the upper bound for generation quality.
    
    \item \textbf{Direct Reuse (ZERO):} A naive baseline that directly grafts previously computed KV states without any rectification, defining the theoretical upper bound for efficiency. This typically suffers from accuracy degradation due to unaddressed deviation patterns.
\end{itemize}

\textbf{KV Reuse Methods:}
\begin{itemize}
    \item \textbf{CacheBlend}~\cite{yao2025cacheblend}: A selective recomputation method that identifies high-deviation tokens based on value norm deviation in second layer. We set the recomputation token size $K$ dynamically to the top $\alpha=20\%$ of the relay segment length ($K = \lfloor N_{\text{relay}} \times 0.2 \rfloor$).  In long-context scenarios, this linear scaling overhead ($K \propto N$) significantly limits the end-to-end speedup compared to RelayCaching's sparse selection strategy.
    
    \item \textbf{EPIC}~\cite{hu2024epic}: A position-independent caching method that recomputes only a fixed set of prefix tokens. We implement the LegoLink algorithm with a chunk size of 16. However, because EPIC only recomputes a small fixed window of initial tokens at each chunk boundary, it cannot adapt to the distribution shifts that prefix changes induce in the remaining cached context, which in our experiments leads to notable accuracy degradation on reasoning tasks (Table~\ref{tab:main_results_refined}).
    
    \item \textbf{KVCOMM}~\cite{ye2025kvcomm}: A retrieval-based framework that approximates cross-context KV states via similarity matching with historical anchors. We use the default configuration from the official repository with an anchor pool size of 20 per agent role. During inference, the system retrieves a subset of nearest anchors (weighted by softmax similarity) to estimate KV offsets. While robust in many scenarios, its rigid matching constraints cause the reuse rate to decline as the number of agents increases.
\end{itemize}

\subsection{Main Evaluation Benchmarks}
\label{app:benchmarks}
\textbf{Tasks and Prompts.}
\begin{itemize}
    \item \textbf{GSM8K (Reasoning):} A dataset of high-quality grade school math problems. Evaluated in a 4-agent chain structure where agents sequentially verify and refine mathematical reasoning steps.
    \item \textbf{HumanEval (Coding):} A benchmark comprising 164 hand-written Python programming problems. Evaluated in a Developer-Tester loop where the Developer generates code and the Tester provides unit test feedback. The maximum generation length is set to 512 tokens.
    \item \textbf{MMLU (Knowledge):} A massive multitask benchmark covering 57 subjects across STEM, the humanities, and social sciences. Evaluated using a RAG pipeline (Retriever-Reader-Critic) to test the relay of long contexts containing external knowledge.
    \item \textbf{AIME (Efficiency):} Used strictly for efficiency scaling laws (Figure~\ref{fig:ttft_scaling}). We constructed synthetic prompts with controlled input lengths ranging from 512 to 12,288 tokens. The outputs were aligned to ensure fair comparison of TTFT across context lengths.
\end{itemize}

\textbf{Hardware and Environment.}
All experiments were conducted on a server equipped with NVIDIA H20 (96GB) GPUs. The system is implemented on top of PyTorch v2.1.0 and Transformers v4.50.2. We measure TTFT using CUDA Events for precise latency recording.

\subsection{Detailed Agent Configurations}
\label{app:agent_configs}

In this section, we provide the exact agent roles, system prompts, and interaction topologies employed in our multi-agent experiments.

\textbf{GSM8K (Reasoning).}
We utilize a sequential chain structure comprising three distinct agents followed by a decision node. The workflow proceeds as \texttt{Mathematical Analyst} $\rightarrow$ \texttt{Inspector} $\rightarrow$ \texttt{Math Solver} $\rightarrow$ \texttt{Decision Maker}.
\begin{itemize}
    \item \textbf{Agent 1: Mathematical Analyst.}
    \begin{quote}
    \textit{"You are a mathematical analyst. You will be given a math problem, analysis and code from other agents. You need to first analyze the problem-solving process step by step, where the variables are represented by letters. Then you substitute the values into the analysis process to perform calculations and get the results. The last line of your output contains only the final result without any units, for example: The answer is 140. You will be given some examples you may refer to."}
    \end{quote}
    
    \item \textbf{Agent 2: Inspector.}
    \begin{quote}
    \textit{"You are an Inspector. You will be given a math problem, analysis and code from other agents. Check whether the logic/calculation of the problem solving and analysis process is correct(if present). Check whether the code corresponds to the solution analysis(if present). Give your own solving process step by step based on hints. The last line of your output contains only the final result without any units, for example: The answer is 140. You will be given some examples you may refer to."}
    \end{quote}

    \item \textbf{Agent 3: Math Solver.}
    \begin{quote}
    \textit{"You are a math expert. You will be given a math problem and hints from other agents. Give your own solving process step by step based on hints. The last line of your output contains only the final result without any units, for example: The answer is 140."}
    \end{quote}

    \item \textbf{Decision Node: Decision Maker.}
    \begin{quote}
    \textit{"You are a decision-maker. You check if the solutions match the question. You prefer simple and direct answers. You will be given a math problem (Q) and solutions from other agents. Read the Question (Q) carefully. It is the absolute truth. If an agent's solution uses the numbers from Q and the logic is simple and correct, ACCEPT it directly. Do NOT add extra steps or change the conditions in Q. Only provide a new solution if the agents made a calculation error. The last line of your output contains only the final result without any units, for example: The answer is 140"}
    \end{quote}
\end{itemize}

\textbf{HumanEval (Coding).}
Our concrete implementation instantiates the conceptual Developer-Tester loop described in the main text as a robust 5-stage pipeline, ensuring rigorous quality control before the final decision aggregation. The execution flow is \texttt{Project Manager} $\rightarrow$ \texttt{Algorithm Designer} $\rightarrow$ \texttt{Programming Expert} $\rightarrow$ \texttt{Test Analyst} $\rightarrow$ \texttt{Bug Fixer} $\rightarrow$ \texttt{Top Decision-Maker}.
\begin{itemize}
    \item \textbf{Agent 1: Project Manager.}
    \begin{quote}
    \textit{"You are a project manager. You will be given a function signature and its docstring by the user. You are responsible for overseeing the overall structure of the code, ensuring that the code is structured to complete the task. Implement code concisely and correctly without pursuing over-engineering. You need to suggest optimal design patterns to ensure that the code follows best practices for maintainability and flexibility. You can specify the overall design of the code, including the classes that need to be defined(maybe none) and the functions used."}
    \end{quote}
    
    \item \textbf{Agent 2: Algorithm Designer.}
    \begin{quote}
    \textit{"You are an algorithm designer. You will be given a function signature and its docstring by the user. You need to specify the specific design of the algorithm, including the classes that may be defined and the functions used. You need to generate the detailed documentation, including explanations of the algorithm, usage instructions, and API references. When the implementation logic is complex, you can give the pseudocode logic of the main algorithm. I hope your reply will be more concise. Preferably within fifty words. Don’t list too many points."}
    \end{quote}

    \item \textbf{Agent 3: Programming Expert (Developer).}
    \begin{quote}
    \textit{"You are a programming expert. You will be given a function signature and its docstring by the user. You may be able to get the output results of other agents. They may have passed internal tests, but they may not be completely correct. Write your full implementation (restate the function signature). Use a Python code block to write your response. For example: \texttt{print('Hello world!')} Do not include anything other than Python code blocks in your response. Do not change function names and input variable types in tasks."}
    \end{quote}

    \item \textbf{Agent 4: Test Analyst (Tester).}
    \begin{quote}
    \textit{"You are a test analyst. You will be given a function signature and its docstring by the user. You need to provide problems in the current code or solution based on the test data and possible test feedback in the question. You need to provide additional special use cases, boundary conditions, etc. that should be paid attention to when writing code. You can point out any potential errors in the code."}
    \end{quote}

    \item \textbf{Agent 5: Bug Fixer.}
    \begin{quote}
    \textit{"You are a bug fixer. You will be given a function signature and its docstring by the user. You need to provide modified and improved python code based on the current overall code. Use a Python code block to write your response."}
    \end{quote}

    \item \textbf{Decision Node: Top Decision-Maker.}
    \begin{quote}
    \textit{"You are the top decision-maker and are good at analyzing and summarizing other people's opinions, finding errors and giving final answers. And you are an AI that only responds with only python code. You will be given a function signature and its docstring by the user. You may be given the overall code design, algorithm framework, code implementation or test problems. Write your full implementation (restate the function signature). If the prompt given to you contains code that passed internal testing, you can choose the most reliable reply. If there is no code that has passed internal testing in the prompt, you can change it yourself according to the prompt. Use a Python code block to write your response... Do not include anything other than Python code blocks in your response"}
    \end{quote}
\end{itemize}

\textbf{MMLU (Knowledge).}
To mitigate hallucinations, we employ a diverse five-agent group followed by an expert analyzer for final selection. The roles rotate through: \texttt{Knowledgeable Expert} $\rightarrow$ \texttt{Wiki Searcher} $\rightarrow$ \texttt{Critic} $\rightarrow$ \texttt{Mathematician} $\rightarrow$ \texttt{Psychologist} $\rightarrow$ \texttt{Expert Analyzer}.
\begin{itemize}
    \item \textbf{Agent 1: Knowledgeable Expert.}
    \begin{quote}
    \textit{"Please give at most six key entities that need to be searched in wikipedia to solve the problem. Key entities that need to be searched are included between two '@' when output... If there is no entity in the question that needs to be searched in Wikipedia, you don't have to provide it."}
    \end{quote}
    
    \item \textbf{Agent 2: Wiki Searcher.}
    \begin{quote}
    \textit{"Please refer to them step by step to give your answer. If the Wikipedia overview is missing or insufficient, you must rely on your internal knowledge to explicitly define the nature, social status, or scientific mechanism of the entities identified by Agent 1 BEFORE choosing an option. Do not guess based on vague associations."}
    \end{quote}

    \item \textbf{Agent 3: Critic.}
    \begin{quote}
    \textit{"You are an excellent critic. You will review the analysis provided by the Wiki Searcher. Please critique the analysis point by point based on the following criteria: ... 3. Is the final option fully supported by the analysis steps?"}
    \end{quote}
    
    \item \textbf{Agent 4: Mathematician.}
    \begin{quote}
    \textit{"You are a mathematician who is good at math games, arithmetic calculation, and long-term planning."}
    \end{quote}
    
    \item \textbf{Agent 5: Psychologist.}
    \begin{quote}
    \textit{"You are a psychologist. You are good at psychology, sociology, and philosophy. You give people scientific suggestions that will make them feel better."}
    \end{quote}

    \item \textbf{Decision Node: Expert Analyzer.}
    \begin{quote}
    \textit{"You are an expert analyzer. You are good at finding the correct option among A, B, C and D based on logic and evidence. You will be given a question with 4 options (A, B, C, D) and analysis from other agents. If the analysis from other agents is logical and correct, use it to support your answer. If the analysis from other agents is wrong, ignore it and use your own reasoning. Do not imitate the other agents' tone; just state the facts. Write a brief analysis. The last line of your output must contain only the option letter, for example: The answer is A"}
    \end{quote}
\end{itemize}

\textbf{AIME (Efficiency Evaluation).}
For the efficiency scaling laws analyzed in Figure~\ref{fig:ttft_scaling} and Table~\ref{tab:ttft_breakdown}, we employ a controlled experimental setup to decouple system latency from model generation variance using the AIME 2024 dataset.

\begin{itemize}
    \item \textbf{Dataset Processing:} We construct the task input using a fixed template:
    \begin{quote}
    \textit{"You are given a math competition problem. \\
    Problem: \{problem\} \\
    Task: \\
    - Solve the problem carefully. \\
    - Explain your reasoning step by step. \\
    - Finish by stating ONLY the final answer on the last line."}
    \end{quote}
    To ensure precise control over prefill and decode lengths, the input text is padded with a neutral instruction set to exactly match the target token count:
    \begin{quote}
    \textit{"Additional instructions: \\
    - Use clear algebra/number theory steps. \\
    - Check constraints and edge cases. \\
    - Provide the final answer on the last line."}
    \end{quote}
    
    \item \textbf{Agent Configuration:} We utilize a homogeneous agent structure configured to form a strict cumulative context chain. The system prompt is set to empty. Agent $i$ receives a user prompt consisting of the direct concatenation without separators: \texttt{\{user\_question\}\{output\_agent\_0\}...\{output\_agent\_(i-1)\}}. This guarantees that the prefill workload for downstream agents scales precisely as defined in the experiment.
\end{itemize}

\section{Discussion}

\paragraph{Extension to Heterogeneous Settings.}
Our current framework focuses on homogeneous multi-agent systems where all agents share identical model architectures. In heterogeneous settings—where agents may use different model families or sizes—direct KV cache transfer becomes infeasible due to mismatched hidden dimensions and attention head configurations. Potential directions include learning lightweight projection layers between architectures or developing distillation-based approaches that transfer semantic information rather than raw KV caches. We leave this exploration to future work.
\paragraph{Hyperparameter Sensitivity.}
RelayCaching relies on two types of hyperparameters: 
(1) layer-range boundaries ($L_{\text{start}}$, $L_{\text{end}}$, $L_{\text{det}}$) determined through offline profiling on a calibration dataset, and 
(2) threshold multipliers ($\tau_{\text{dev}}$, $\tau_{\text{inf}}$) that are fixed empirical constants applied to runtime-computed statistics ($\mu_{\text{dev}}$, $\mu_{\text{inf}}$).
While both generalize well across our evaluated benchmarks, 
task distributions with substantially different characteristics 
may benefit from task-specific calibration or online adaptation. 
%%%%%%%%%%%%%%%%%%%%%%%%%%%%%%%%%%%%%%%%%%%%%%%%%%%%%%%%%%%%%%%%%%%%%%%%%%%%%%%
%%%%%%%%%%%%%%%%%%%%%%%%%%%%%%%%%%%%%%%%%%%%%%%%%%%%%%%%%%%%%%%%%%%%%%%%%%%%%%%
\end{document}

%% file: contents/1-intro.tex
\vskip -0.1in
\section{Introduction}
\begin{figure*}[t]
  \centering
  \includegraphics[width=\textwidth]{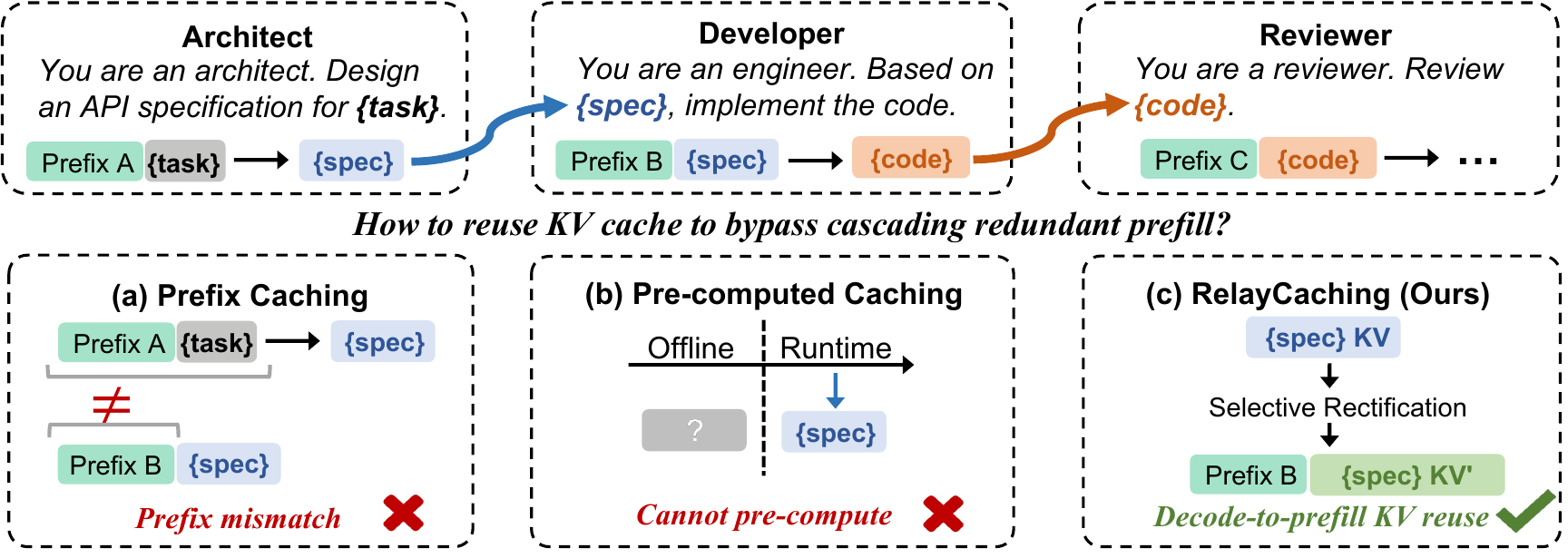}
  \setlength{\abovecaptionskip}{-5pt}
  
\setlength{\belowcaptionskip}{-10pt}
  \caption{
  Overview of KV cache reuse in multi-agent LLM collaboration.
  \textbf{Top:} A typical workflow where an Architect generates 
  \texttt{\{spec\}}, a Developer produces \texttt{\{code\}} based on it, 
  and a Reviewer evaluates the result, so each agent's output becomes 
  the next agent's input, causing cascading redundant prefill.
  \textbf{Bottom:} Comparison of caching strategies. (a) Prefix Caching 
  fails due to prefix mismatch; (b) Pre-computed Caching cannot handle 
  dynamic content; (c) RelayCaching enables decode-to-prefill 
  KV reuse via selective rectification.
} 
  \label{fig:reuse-comparison}
\end{figure*}

Multi-agent systems (MAS) have emerged as a dominant paradigm for deploying LLMs on complex tasks, orchestrating specialized agents that collaborate through structured communication~\cite{wu2024autogen,hong2023metagpt,li2023camel}.
Recent frameworks scale MAS to increasingly sophisticated workflows—from multi-turn software development pipelines~\cite{qian2024chatdev} to scientific discovery conducted by teams of coordinated agents~\cite{su2025many}—thereby amplifying both capability and computational demands.
In these pipelines, agents exchange intermediate outputs such as reasoning chains, API or code artifacts, and summarized context, enabling task decomposition and iterative refinement~\cite{wu2024autogen,hong2023metagpt}.
As illustrated in Figure~\ref{fig:reuse-comparison}, consider a development workflow:  an Architect agent designs an API specification, a Developer implements it,  and a Reviewer evaluates the result.
Each upstream output is embedded into downstream prompts, causing cascading redundant prefill. 

However, this output-to-input data flow suffers from \textit{prefix variation}: 
the same tokens appear with different preceding contexts between decoding and prefill phases.
This mismatch introduces severe \textit{cascading context redundancy}, 
as serving systems must treat previously generated text as entirely new input 
and recompute KV caches from scratch during prefill.
In a pipeline with $M$ interaction turns, each agent must reprocess the accumulated history 
from all preceding interactions, causing cumulative prefill cost quadratically with interaction turns ($O(M^2)$)~\cite{ye2025kvcomm}. 
This redundant prefill computation for shared content becomes a critical bottleneck 
in multi-agent pipelines, significantly increasing KV-cache memory usage and TTFT. 

Existing prefill redundancy optimization methods based on \textit{prefix caching} and \textit{pre-computed caching} are inadequate for the dynamic nature of MAS.
\textit{Prefix caching}~\cite{kwon2023efficient, zheng2024sglang} relies on strict positional alignment and fails in MAS, where reused content frequently appears at non-prefix positions.
\textit{Pre-computed caching}~\cite{gim2024prompt, yao2025cacheblend} enables non-prefix reuse but assumes static, offline content (e.g., documents in RAG), limiting applicability in MAS where outputs are generated on the fly.
Consequently, current systems often fall back to expensive full prefilling in MAS settings (Figure~\ref{fig:reuse-comparison}(a,b)). 

This limitation motivates a fundamental question: \textit{can we directly reuse decoding KV caches to bypass redundant prefill computation despite prefix variation?}
To answer this, we conduct a systematic empirical study comparing decoding KV caches with their full-prefill counterparts.
Our analysis reveals three key findings that establish the feasibility of direct reuse with targeted rectification:
\textbf{(1) High macro-level alignment.} 
Despite prefix variation, decoding KV caches maintain high alignment with full-prefill counterparts, 
with value cosine similarity emerging as the primary deviation indicator.
\textbf{(2) U-shaped layer-wise similarity profile.}
Middle layers show the largest deviations and dominate subsequent generation quality, 
while shallow and deep layers remain relatively stable. 
\textbf{(3) Sparse and correlated token-wise deviations.} 
Token-wise deviations are sparse and exhibit strong inter-layer correlation. 

Building on these observations, we propose \textit{RelayCaching}, 
a training-free method that directly reuses decoding KV caches 
from previous agents in  prefill phases.
RelayCaching comprises two core components:
(1) a \textit{layer-range profiler} that identifies the critical layer range using the U-shaped profile and selects a detection layer based on inter-layer correlation, and (2) a \textit{token selector} that combines deviation-based and influence-based selection to pinpoint tokens requiring rectification. This design enables RelayCaching to maintain accuracy comparable to full prefilling while rectifying only a small fraction of KV caches, yielding a superior accuracy–efficiency trade-off over existing methods.

In summary, this paper makes the following contributions:
\begin{itemize} 
\setlength{\itemsep}{0pt}
\setlength{\parskip}{0pt}
\setlength{\parsep}{0pt}
\item 
We systematically characterize KV deviations between decoding and prefill phases, 
revealing that decoding KV caches exhibit high alignment with full-prefill counterparts despite prefix variation. 
Residual deviations follow systematic patterns that enable targeted rectification: 
U-shaped similarity across layers and sparse, inter-layer-correlated deviations across tokens. 

\item 
We propose RelayCaching, a training-free method that directly reuses decoding KV caches 
from upstream agents in downstream prefill phases. 
RelayCaching employs layer-range profiling to confine rectification to middle layers 
and token selection to identify critical tokens, 
achieving efficient cache alignment with minimal overhead.

\item 
We demonstrate that RelayCaching maintains generation quality comparable to full prefilling 
across diverse tasks while achieving over 80\% KV cache reuse 
and up to 4.7$\times$ TTFT reduction in multi-agent systems.
\end{itemize} 

%% file: contents/2-background.tex
%%%%GYS Version 0
\section{Related Works}
\subsection{LLM-Based Multi-Agent Workflows}
\label{subsec:mas}
LLM-based Multi-Agent Systems have emerged as a prevalent paradigm for solving complex tasks through collective intelligence. Foundational frameworks such as CAMEL~\cite{li2023camel} and AutoGen~\cite{wu2024autogen} pioneered role-playing for task decomposition, while MetaGPT~\cite{hong2023metagpt} introduces Standard Operating Procedures for assembly-line workflows. Recent frameworks like GPTSwarm~\cite{zhuge2024gptswarm} and MegaAgent~\cite{wang2025megaagent} further model agent collectives as dynamic computation graphs.
However, these pipelines suffer from \textit{cascading context redundancy}, where upstream outputs are repeatedly reused under changing prefixes, forcing KV cache rebuilding and incurring significant overhead.

\subsection{Efficient LLM Inference and KV Caching}
\label{subsec:kv_caching}
KV caching has become the de facto standard for mitigating redundant prefill computation. We categorize existing approaches into prefix caching and pre-computed caching.
Prefix caching reuses KV caches for recurring, identical prefixes. Systems like vLLM~\cite{kwon2023efficient} and SGLang~\cite{zheng2024sglang} introduce PagedAttention and RadixAttention to share prefix KV caches across requests. 
Pre-computed caching relaxes the strict prefix constraint by encoding reusable contexts in advance.
Prompt Cache~\cite{gim2024prompt} enables reuse for frequent prompt modules; Block-Attention~\cite{mablock} reuses KV caches for static documents with fine-tuning.
KVLink~\cite{yang2025kvlink} and APE~\cite{yangape} align independently encoded KV caches via trainable tokens or adaptive scaling; CacheBlend~\cite{yao2025cacheblend} and EPIC~\cite{huepic} correct precomputed caches through dynamic or static token recomputation; DroidSpeak~\cite{liu2024droidspeak} extends pre-computed caching across models with shared architectures.
Despite these advances, prefix caching requires strict positional alignment 
that agent-generated contexts rarely satisfy, 
while pre-computed caching assumes static, offline-encodable content—an assumption violated by dynamically generated outputs, cannot be directly applied in multi-agent settings.
These limitations motivate \textit{decode-to-prefill KV reuse}, which transfers KV caches from one agent's decoding phase 
to another's prefilling phase.
Unlike pre-computed caching, this paradigm enables targeted rectification of runtime-generated KV caches without prefix alignment constraints, preserving accuracy with minimal overhead.

\subsection{Efficiency in Collaborative Inference} 
\label{subsec:collab_efficiency}
Recent efforts to accelerate collaborative inference focus on two orthogonal aspects: topology sparsification and state reuse.
Topology sparsification aims to reduces the volume of agent interactions. AgentPrune~\cite{zhangcut} and AgentDropout~\cite{wang-etal-2025-agentdropout} optimize message-passing graphs by pruning redundant agents, but treat model inference as a black box.
State reuse seeks to minimizes computational cost per interaction. To the best of our knowledge, KVCOMM~\cite{ye2025kvcomm} is the first to explicitly address \textit{cascading context redundancy} via retrieval-based anchor pools, but relies on external samples and suffers from degraded reuse rates at scale. RelayCaching instead directly exploits the inherent similarity between decoding and 
prefill KV caches, enabling fine-grained rectification at layer and token granularity without external reference. 

%% file: contents/3-observation.tex
\section{Observations}
\label{sec:observation}
In this section, we investigate the alignment and deviation 
between prefill and decoding KV caches under \emph{decode-to-prefill KV reuse}, where the same tokens appear under different preceding contexts.
We address two questions: 
(1)~\emph{Reusability}: are decoding KV caches sufficiently aligned 
with full-prefill counterparts despite prefix variation? 
(2)~\emph{Deviation patterns}: do residual deviations exhibit systematic patterns that can guide efficient rectification?
We conduct experiments on 2WikiMQA~\cite{ho2020constructing} with Mistral-7B-Instruct-v0.3~\cite{jiang2023mistral7b} using a summarize-then-answer workflow.
We further verify that our findings generalize to other model architectures in the
Appendix~\ref{app:further_observations}.

\subsection{Macro-Level KV Similarity}
\label{subsec:obs_macro}
\begin{figure}[t]
  \centering
  \includegraphics[width=0.9\linewidth]{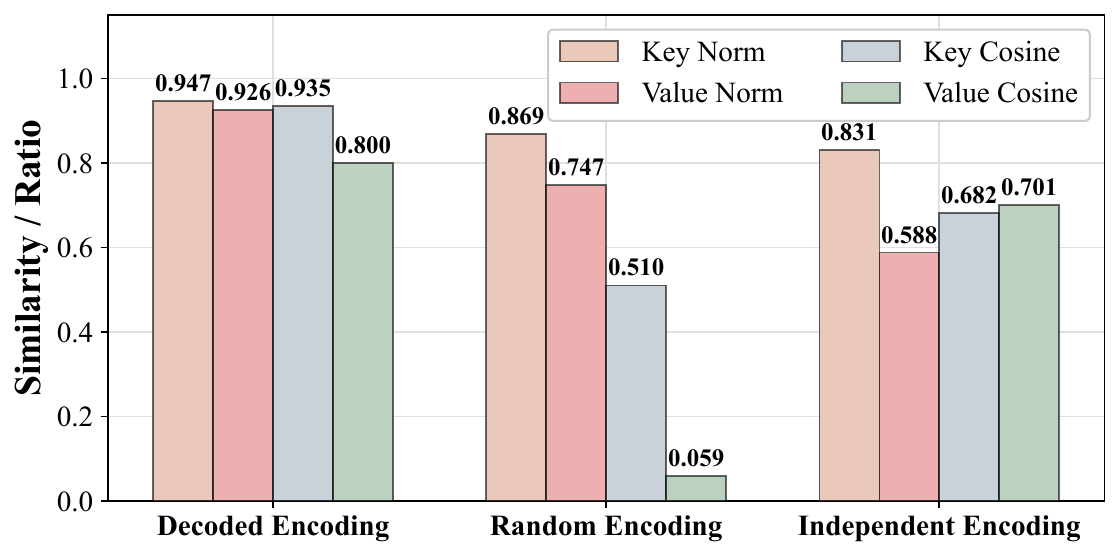}
  \setlength{\abovecaptionskip}{5pt} % 默认通常是 10pt 左右，设小一点
\setlength{\belowcaptionskip}{-18pt} % 标题下方的距离也可以调
  \caption{\textbf{Macro-level KV similarity.}
Average cosine and norm similarity between \emph{Decoding KV} caches and \emph{Full-Prefill KV} caches across layers and tokens on 2WikiMQA, compared against \emph{Random KV} and \emph{Independent KV}.}
  \label{fig:macro_sim}
\end{figure}
We first validate that decoding KV caches remain globally aligned with full-prefill counterparts, 
finding that \emph{value cosine similarity} emerges as the primary deviation signal.
We compare cosine and norm similarity under three settings: 
(1)~\emph{Decoding KV}, produced during decoding with shifted prefixes; 
(2)~\emph{Random KV}, from prefill with the summary replaced by random tokens; 
and (3)~\emph{Independent KV}, from prefill on summary tokens alone without preceding prefix.
As shown in Figure~\ref{fig:macro_sim}, \emph{Decoding KV} maintains high macro-level similarity, 
whereas \emph{Random KV} shows much lower similarity with value cosine approaching zero.
Notably, keys in \emph{Decoding KV} align closely with full-prefill keys in both direction and magnitude, 
whereas values show directional perturbations while largely preserving their magnitudes.
In contrast, \emph{Independent KV} exhibits a distinct pattern: 
value norms deviate substantially from full-prefill counterparts, 
indicating that pre-computation schemes relying on independently encoded KV caches 
are incompatible with \emph{decode-to-prefill KV reuse}.

\textbf{Insights.} 
Decoding KV caches remain highly aligned with full-prefill counterparts, validating the reuse premise.
Value cosine similarity is the lowest among all metrics, so we focus on it in subsequent layer- and token-wise analysis.

\begin{figure}[h]
  \centering
  \begin{subfigure}{\linewidth}
    \centering
     \includegraphics[width=0.9\linewidth]{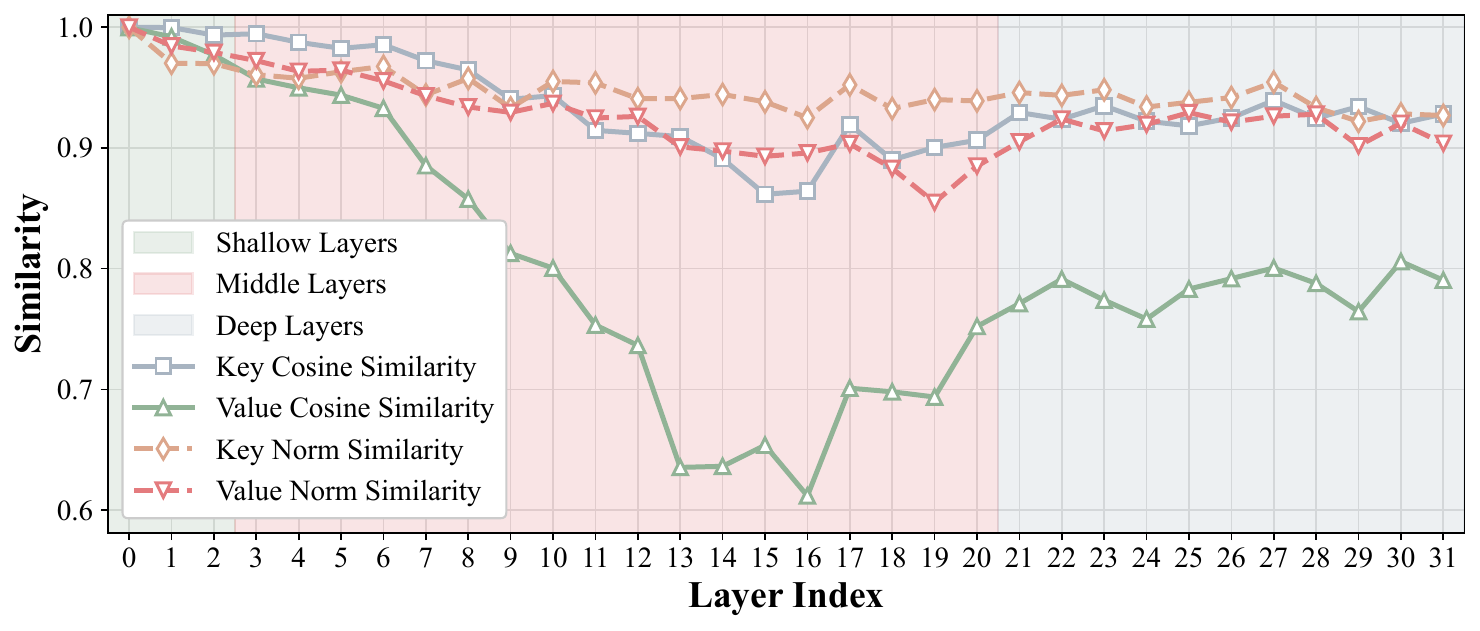}
     \vskip -0.5em
     \caption{}
    \label{fig:layer_curve_raw}
  \end{subfigure}

  \begin{subfigure}{\linewidth}
    \centering
    \includegraphics[width=0.9\linewidth]{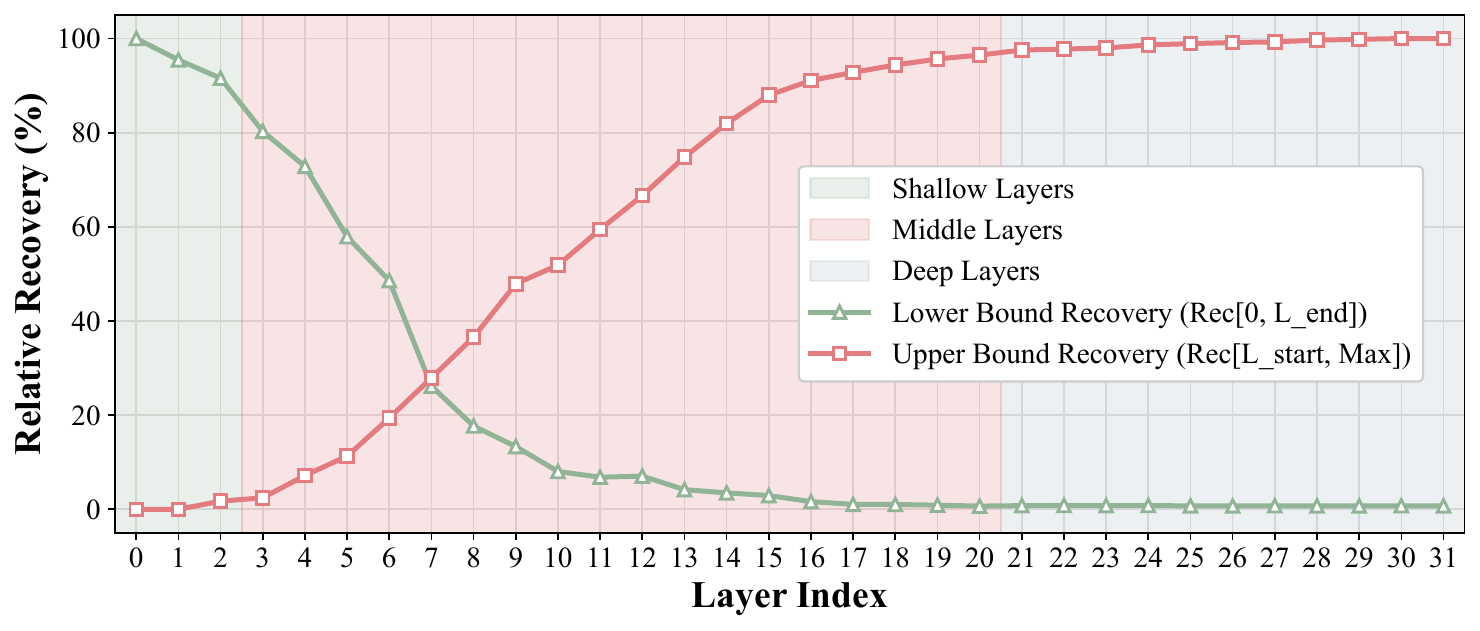}
    \vskip -0.5em
    \caption{}
    \label{fig:layer_curve_recompute}
  \end{subfigure}
\setlength{\abovecaptionskip}{-10pt} 
\setlength{\belowcaptionskip}{-18pt}
 \caption{\textbf{Layer-wise structured deviations.} 
 % cosine similarity and norm similarity
\textit{(a)} Layer-wise similarity between decoding and full-prefill KV caches, highlighting a U-shaped value cosine similarity profile across layers. 
\textit{(b)} Relative recovery of value cosine similarity for tokens after the reuse position when decoding KV caches are replaced with their full-prefill counterparts either up to or from a given layer index. 
}
  \label{fig:layer_curve}
\end{figure}
\subsection{Layer-Wise Deviation Pattern}
\label{subsec:obs_layer}
We next analyze how deviations distribute across layers, 
finding that middle layers exhibit the largest deviations and contribute most to subsequent generation errors.
Figure~\ref{fig:layer_curve_raw} reveals a characteristic \emph{U-shaped profile} 
in value cosine similarity: 
high in shallow layers, dropping to a minimum in middle layers, 
and partially recovering in deeper layers.
To verify that middle-layer deviations are the primary error source for subsequent generation, 
we perform an oracle experiment that selectively substitutes decoding KV caches 
with full-prefill KV at different layer ranges.
Figure~\ref{fig:layer_curve_recompute} reports the resulting value cosine similarity 
for tokens following the reused segment under two settings: 
(i)~substituting all layers up to index $i$, and 
(ii)~substituting all layers from $i$ onward.
In both cases, similarity recovery is steepest when the substituted range covers 
the middle layers identified by the U-shaped profile, 
with diminishing returns when extending into shallow or deep layers, confirming that middle layers dominate downstream generation quality.

\textbf{Insights.} 
The U-shaped profile provides a principled criterion for identifying the critical layer range, 
enabling targeted rectification that captures most of the benefit 
while bypassing shallow and deep layers.  
\subsection{Token-Wise Deviation Pattern}
\label{subsec:obs_token}
\begin{figure}[t]
  \centering
  \begin{subfigure}{\linewidth}
    \centering
    \includegraphics[width=0.9\linewidth]{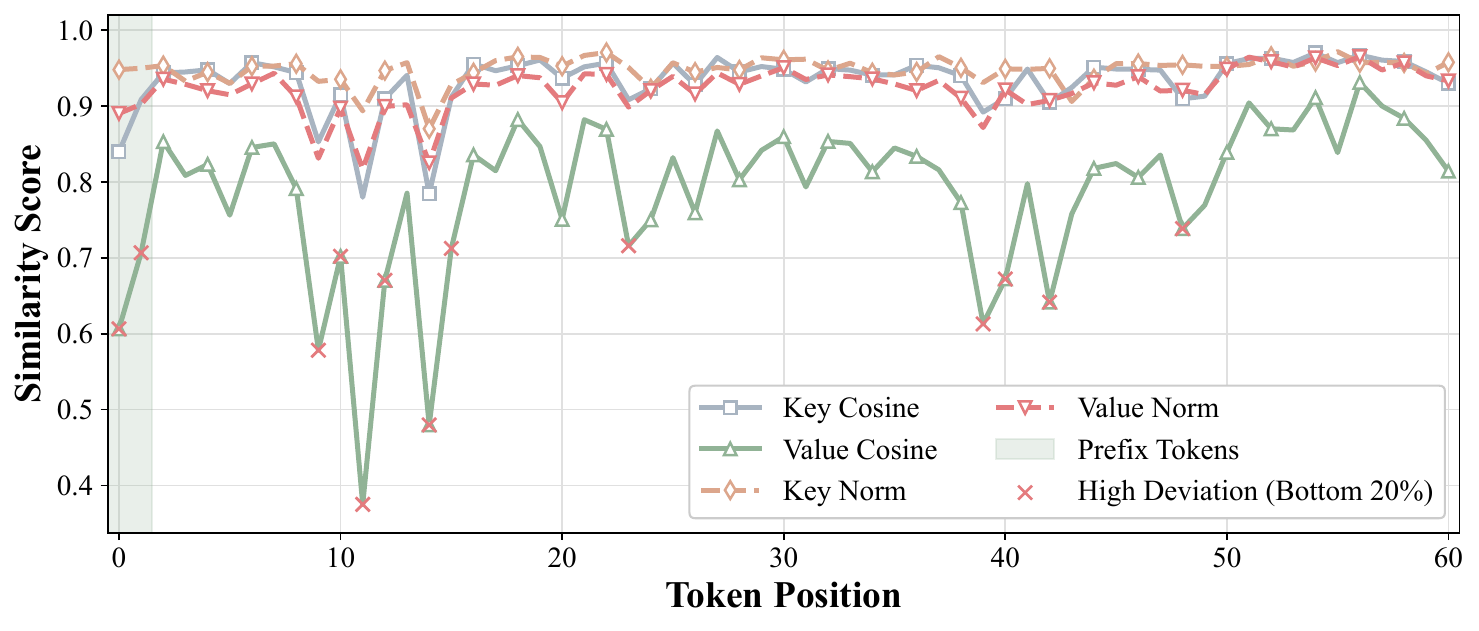}
    \vskip -0.5em
    \caption{}
    \label{fig:token_deviation_profile}
  \end{subfigure}

  \begin{subfigure}{\linewidth}
    \centering
    \includegraphics[width=0.9\linewidth]{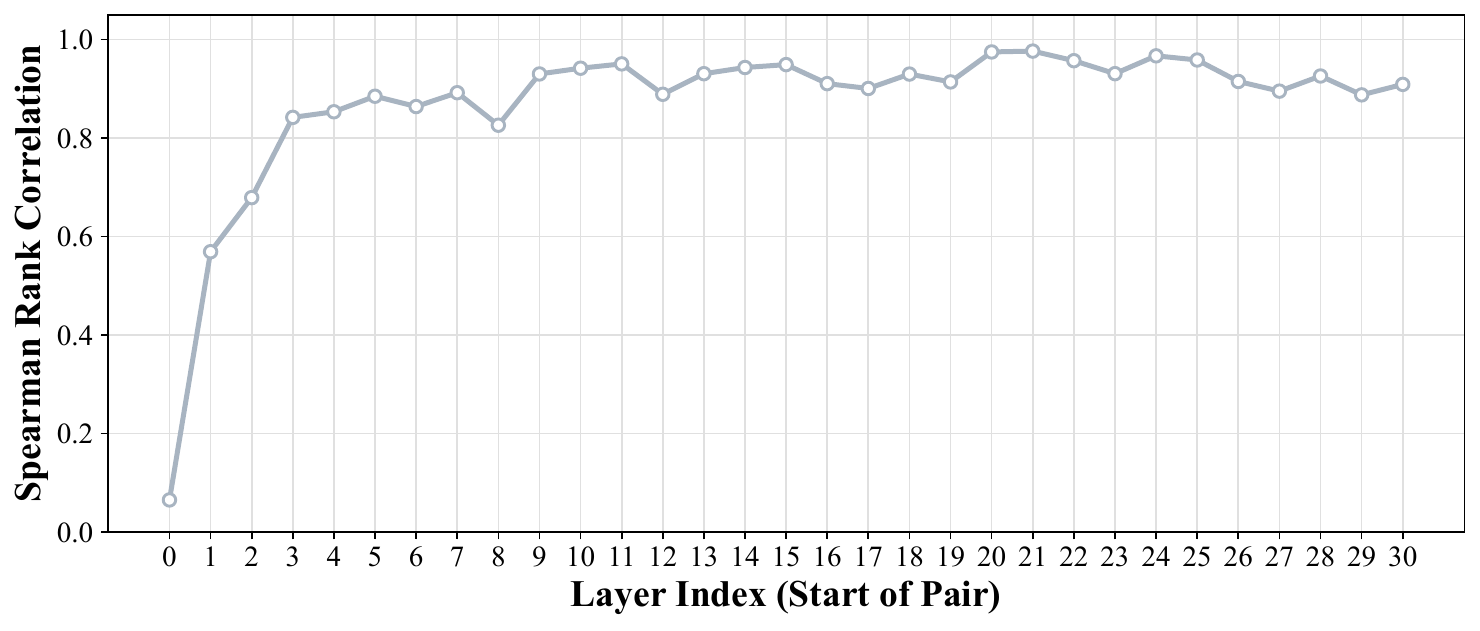}
    \vskip -0.5em
    \caption{}
    \label{fig:token_deviation_correlation}
  \end{subfigure}
\setlength{\abovecaptionskip}{-10pt}
\setlength{\belowcaptionskip}{-18pt} 
  \caption{\textbf{Token-wise structured deviations.}
\textit{(a)} Token-wise similarity between decoding and full-prefill KV caches, averaged over layers, revealing a sparse set of high-deviation positions.
\textit{(b)} Spearman rank correlation of token-wise value cosine deviations between adjacent layers, showing that high-deviation positions, once emerged, tend to persist.}
\label{fig:token_structured_deviation}
\end{figure}

Finally, we investigate how deviations distribute across tokens, finding that token-level deviations are sparse and exhibit strong inter-layer correlation. 
As shown in Figure~\ref{fig:token_deviation_profile}, only a small subset of positions exhibits large deviations, 
while most tokens maintain high similarity with full-prefill counterparts.
Some of these high-deviation tokens are position-dependent, such as tokens near the start of the reuse segment, which are inherently more sensitive 
to prefix variation; others are content-dependent, 
reflecting input-specific sensitivity to the preceding context. 
Figure~\ref{fig:token_deviation_correlation} shows that the Spearman rank correlation 
of token-wise deviations between adjacent layers starts low but rapidly increases and plateaus, indicating that high-deviation positions persist once they emerge.

\textbf{Insights.} 
The combination of token-wise sparsity and inter-layer correlation 
implies that rectification can be highly selective: 
by identifying high-deviation tokens at a layer where rankings have stabilized, 
we can propagate this selection to subsequent layers, 
achieving most of the rectification benefit at minimal cost.

%% file: contents/4-relaycaching.tex
\section{RelayCaching}

\subsection{Overall Architecture}
\label{sec:method}

\begin{figure*}[t]
  \centering
  \includegraphics[width=\textwidth]{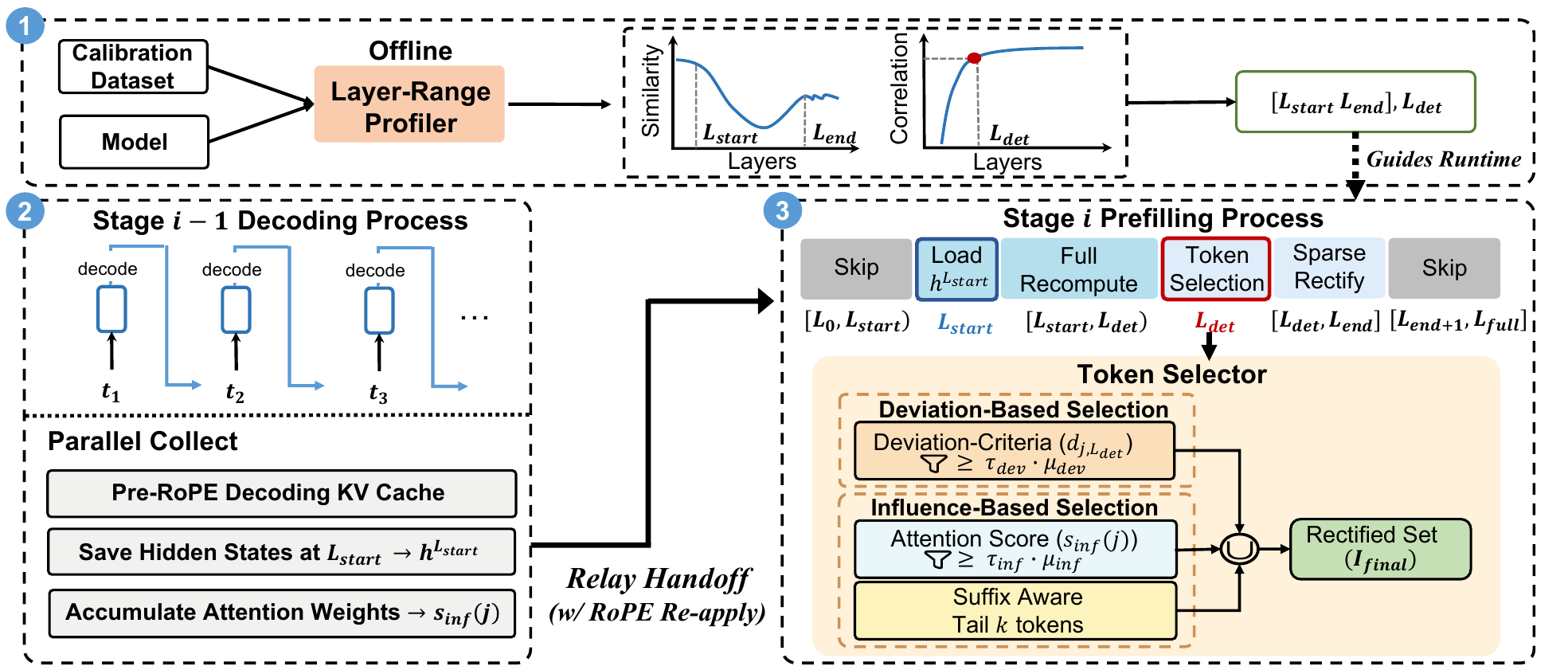}
  \setlength{\abovecaptionskip}{-3pt} 
\setlength{\belowcaptionskip}{-10pt} 
  \caption{\textbf{Overview of RelayCaching.}
(1) An offline layer-range profiler analyzes calibration data to identify a critical layer range $[L_{\text{start}}, L_{\text{end}}]$ from the similarity profile and a detection layer $L_{\text{det}}$ from the inter-layer correlation profile.
(2) During stage-$(i{-}1)$ decoding, the cache manager stores pre-RoPE KV caches, saves hidden states at $L_{\text{start}}$, and accumulates influence scores $s_{\text{inf}}(j)$.
(3) At stage-$i$ prefilling, RelayCaching re-applies RoPE to align reused KV caches to new positions, performs token selection at $L_{\text{det}}$ via combined deviation- and influence-based criteria, and finally applies sparse rectification on selected tokens within $[L_{\text{start}}, L_{\text{end}}]$.}
  \label{fig:relay_arch}
\end{figure*}

Guided by Section~\ref{sec:observation}, we present RelayCaching, a training-free method that accelerates multi-agent collaboration by reusing decoding KV caches with targeted rectification.
The key insight is that deviations concentrate in middle layers and a sparse set of tokens, enabling accurate repair at these position instead of full-prefill recomputation.
As depicted in Figure~\ref{fig:relay_arch}, RelayCaching comprises two components:
a \textbf{layer-range profiler} that exploits the U-shaped similarity profile and inter-layer correlation identifies the critical layer range $[L_{\text{start}}, L_{\text{end}}]$ 
and a detection layer $L_{\text{det}}$;
and a \textbf{token selector} that combines deviation-based and influence-based criteria. 
Together, these enable selective recomputation of identified tokens within the critical range, preserving generation quality at a fraction of the cost.

\subsection{Layer-Range Profiler}

The layer-range profiler determines \emph{where} to apply rectification and \emph{at which layer} to perform token selection.
The U-shaped profile (Section~\ref{subsec:obs_layer}) suggests rectification should concentrate on middle layers $[L_{\text{start}}, L_{\text{end}}]$, while inter-layer correlation (Section~\ref{subsec:obs_token}) enables identifying a detection layer $L_{\text{det}}$ for reliable token selection.
The profiler performs offline profiling on auxiliary datasets to determine these parameters.
Let $\mathbf{I}_{\text{reuse}}$ denote the set of token positions in the reuse segment.
To quantify the alignment between reused and full-prefill KV caches, 
we define the token-wise deviation at position $j$ and layer $\ell$ as
\begin{equation}
d_{j,\ell} = 1 - \frac{1}{H} \sum_{h=1}^{H} \cos(\mathbf{v}_{h,j,\ell}^{\text{reuse}}, \mathbf{v}_{h,j,\ell}^{\text{full}}),
\end{equation}
where $H$ is the number of heads, 
and $\mathbf{v}_{h,j,\ell}^{\text{reuse}}$, $\mathbf{v}_{h,j,\ell}^{\text{full}}$ denote 
the value vectors of head $h$ at position $j$ in layer $\ell$ 
under reused decoding and full-prefill, respectively.
The layer-wise similarity is then the average over all reused tokens:
\begin{equation}
   s_\ell = \frac{1}{|\mathbf{I}_{\text{reuse}}|} \sum_{j \in \mathbf{I}_{\text{reuse}}} (1 - d_{j,\ell}). 
\end{equation}

\noindent\textbf{Start layer.}
Since shallow layers already exhibit high similarity and require no rectification, 
they can provide a high-fidelity starting point for subsequent computation.
We set $L_{\text{start}}$ to the deepest layer with $s_\ell \ge \tau_{\text{st}}$.
At inference time, we load cached hidden states $\mathbf{h}^{(L_{\text{start}})}$ as the starting point for rectification rather than recomputing shallow layers.

\noindent\textbf{End layer.}
Deeper layers  enter a stable regime where deviations partially recover and additional rectification yields diminishing returns on output quality.
We locate the onset of this regime by scanning forward from the minimum-similarity layer $\ell_{\text{min}} = \arg\min_\ell s_\ell$.
We estimate a baseline $b = \mu_{\text{last}} - \sigma_{\text{last}}$, 
where $\mu_{\text{last}}$ and $\sigma_{\text{last}}$ are the mean and standard deviation of $s_\ell$ over the last $T$ layers.
$L_{\text{end}}$ is set to the first layer followed by $C$ consecutive layers 
satisfying 
\begin{equation}
  s_\ell \ge b \;\text{ and }\; |s_\ell - s_{\ell-1}| < \lambda \sigma_{\text{last}}.  
\end{equation}

\noindent\textbf{Detection layer.}
While the layer range defines \emph{where} to rectify, we also need to determine \emph{at which layer} to select tokens to rectify.
Detecting tokens at $L_{\text{start}}$ may lack discriminative power due to high similarity in shallow layers.
To address this, we identify a detection layer, $L_{\text{det}}$ within $[L_{\text{start}}, L_{\text{end}}]$ where deviation patterns have stabilized.
We examine the Spearman rank correlation $\rho_\ell$ of token-wise deviations between adjacent layers.
We define the second-order difference of the correlation trend as:
\begin{equation}
  \alpha_\ell = \rho_\ell - 2\rho_{\ell-1} + \rho_{\ell-2}   
\end{equation}
   
We then identify $\ell^*$ as the first layer where $\alpha_\ell$ transitions from positive to negative, indicating that correlation growth has begun to decelerate.
The detection layer is set to $L_{\text{det}} = \ell^* + 1$, where the deviation ranking has fully stabilized.

At inference time, layers outside $[L_{\text{start}}, L_{\text{end}}]$ directly reuse decoding KV caches without rectification.
Within this range, layers in $[L_{\text{start}}, L_{\text{det}})$ are fully recomputed 
starting from cached hidden states at $L_{\text{start}}$ to enable deviation detection, 
while layers in $[L_{\text{det}}, L_{\text{end}}]$ apply sparse token-selective rectification only on selected tokens.

\subsection{Token Selection Module}
Section~\ref{subsec:obs_token} reveals that deviations are \emph{sparse} and \emph{inter-layer correlated}, enabling single-pass token identification at $L_{\text{det}}$ that propagates throughout $[L_{\text{det}}, L_{\text{end}}]$.
Beyond deviation magnitude, a token's importance also depends on its \emph{downstream influence}: moderately deviated tokens can substantially impact outputs if subsequent tokens repeatedly attend to them.
Accordingly, RelayCaching selects tokens using two criteria:
Deviation-based selection targets tokens with large $d_{j, L_{\text{det}}}$;
Influence-based selection prioritizes tokens receiving high attention from subsequent positions.
Their union forms the final rectification set.

\subsubsection{Deviation-Based Selection}
Deviation-based selection targets tokens with large directional mismatches at $L_{\text{det}}$.
Using $s_{\text{dev}}(j) = d_{j, L_{\text{det}}}$ and mean $\mu_{\text{dev}}$, we select:
\begin{equation}
\mathbf{I}_{\text{dev}} = \{ j \in \mathbf{I}_{\text{reuse}} \mid s_{\text{dev}}(j) \geq \tau_{\text{dev}} \cdot \mu_{\text{dev}} \},
\end{equation}
where $\tau_{\text{dev}}$ is a scaling factor.
This mean-relative threshold adapts to deviation magnitude.
Due to inter-layer correlation, this selection is reused throughout $[L_{\text{det}}, L_{\text{end}}]$.
% single-pass 
\subsubsection{Influence-Based Selection}
Deviation-based selection may miss tokens with moderate deviation but substantial downstream influence.
We incorporate influence-based selection to identify such tokens.

\noindent\textbf{Attention-derived importance.}
We measure influence by accumulating attention weights received across all layers and decoding steps:
\begin{equation}
s_{\text{inf}}(j) = \sum_{t,\ell,h} \alpha_{t,\ell,h,j},
\end{equation}
where $\alpha_{t,\ell,h,j}$ is the attention weight at decoding step $t$, layer $\ell$, head $h$.
We select influential tokens via
\begin{equation}
\mathbf{I}_{\text{inf-score}} = \{ j \in \mathbf{I}_{\text{reuse}} \mid s_{\text{inf}}(j) \geq \tau_{\text{inf}} \cdot \mu_{\text{inf}} \},
\end{equation}
where $\mu_{\text{inf}}$ is the mean influence  and $\tau_{\text{inf}}$ is a scaling factor.

\noindent\textbf{Suffix-aware correction.}
Attention-derived importance is inherently biased toward early positions.
Meanwhile, suffix tokens near the reuse segment boundary often exert disproportionate influence 
on newly appended content due to attention locality.
To balance this, we additionally include the last $K_{\text{suf}}$ trailing tokens as $\mathbf{I}_{\text{inf-suffix}}$, yielding:
\begin{equation}
\mathbf{I}_{\text{inf}} = \mathbf{I}_{\text{inf-score}} \cup \mathbf{I}_{\text{inf-suffix}}.
\end{equation}

The final set combines both criteria:
\begin{equation}
\mathbf{I}_{\text{final}} = \mathbf{I}_{\text{dev}} \cup \mathbf{I}_{\text{inf}}.
\end{equation}
This dual-criterion selection maintains a small rectification set while effectively correcting deviations, achieving a favorable accuracy–efficiency trade-off.

%% file: contents/5-evaluation.tex
\section{Evaluation}
This section presents analyses to answer the following research questions: 
\textbf{RQ1}: Can RelayCaching maintain generation quality comparable to full prefilling? 
\textbf{RQ2}: Can RelayCaching improve efficiency for real-world multi-agent tasks? 
\textbf{RQ3}: How does each component contribute to the efficiency-accuracy trade-off? 
\textbf{RQ4}: How sensitive is RelayCaching to key hyperparameters?

\subsection{Experiment Setup}
\textbf{Datasets and Workflows.}
For accuracy evaluation, following KVCOMM~\cite{ye2025kvcomm}, we construct fully connected multi-agent systems on three benchmarks:
GSM8K~\cite{cobbe2021training} (math reasoning),
HumanEval~\cite{chen2021evaluating} (coding),
and MMLU~\cite{hendrycks2020measuring} (knowledge QA).
For efficiency profiling, we use AIME24~\cite{maa_aime_2025} to produce long reasoning traces that stress-test long-context efficiency.
Detailed agent configurations are in Appendix~\ref{app:benchmarks}.

\noindent\textbf{Models and Implementation.}
We employ Llama-3.1-8B-Instruct~\cite{dubey2024llama} for reasoning and knowledge tasks, 
Qwen2.5-Coder-7B-Instruct~\cite{hui2024qwen2} for coding, 
and Qwen3-0.6B~\cite{yang2025qwen3} for efficiency experiments.
The layer-range profiler is calibrated on 2WikiMQA~\cite{ho2020constructing} 
to ensure task-agnostic deviation detection. 
Default hyperparameters: $\tau_{\text{down}}{=}1.45$, $\tau_{\text{diff}}{=}1.5$, $L_{\text{suf}}{=}10$.
All models run in BF16 on a single NVIDIA H20 GPU. 

\textbf{Baselines.}
We compare against two reference baselines: \emph{Full-prefill} (FULL) without reuse, and \emph{Direct Reuse} (ZERO) without rectification.
We further include three state-of-the-art methods: CacheBlend~\cite{yao2025cacheblend} and EPIC~\cite{hu2024epic} for pre-computed KV caching, and KVCOMM~\cite{ye2025kvcomm} for multi-agent reuse.
All methods use recommended settings; see Appendix~\ref{app:baselines} for details.

\noindent\textbf{Metrics.}
We report task-specific accuracy (pass@1 for HumanEval, accuracy for MMLU and GSM8K) and \emph{Reuse Rate}, the fraction of KV entries reused.

\begin{figure*}[t]
    \centering
    \vskip -0.75em  
    \includegraphics[width=0.4\textwidth]{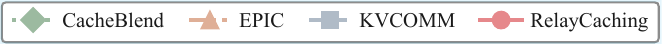}
    \vskip -0.00em
    \captionsetup[subfigure]{skip=1pt, belowskip=-5pt}
    \begin{subfigure}{0.32\textwidth} 
        \centering
        \includegraphics[width=\linewidth]{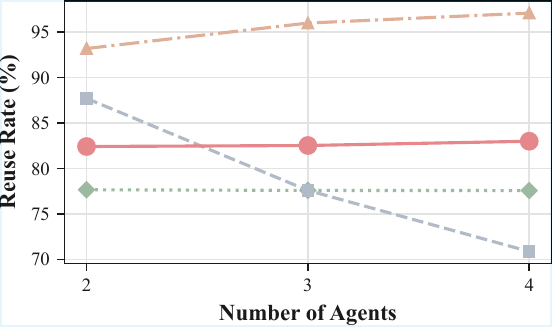}
        \caption{GSM8K} 
        \label{fig:reuse_gsm8k}
    \end{subfigure}
    \hfill 
    \begin{subfigure}{0.32\textwidth}
        \centering
        \includegraphics[width=\linewidth]{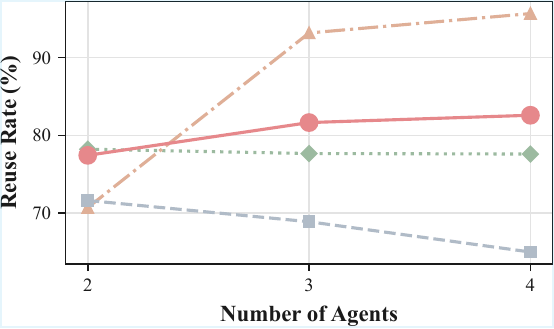}
        \caption{MMLU}
        \label{fig:reuse_mmlu}
    \end{subfigure}
    \hfill
    \begin{subfigure}{0.32\textwidth}
        \centering
        \includegraphics[width=\linewidth]{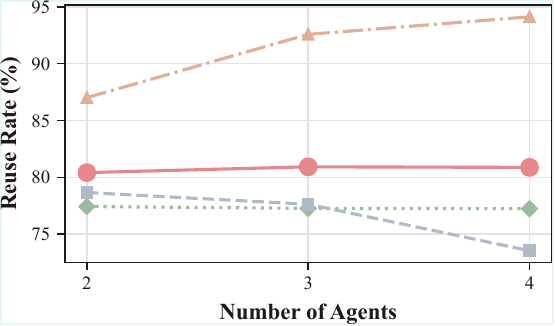}
        \caption{HumanEval}
        \label{fig:reuse_humaneval}
    \end{subfigure}
\setlength{\belowcaptionskip}{-10pt} 
    \caption{\textbf{Computational reuse efficiency across diverse benchmarks.} 
    We compare the Reuse Rate of RelayCaching against baselines on \textit{(a)} GSM8K, \textit{(b)} MMLU, and \textit{(c)} HumanEval as the number of agents increases. }
    \label{fig:reuse_rate_analysis}
\end{figure*}

\subsection{Main Results}
\label{subsec:overall_performance}

\begin{table}[t]
\caption{\textbf{RelayCaching consistently maintains performance comparable to the \textit{Full-prefill}.} Columns indicate the number of agents in the workflow. \textbf{Bold} indicates the best performance; \underline{underline} marks the second best.}
\setlength{\belowcaptionskip}{-10pt} 
\label{tab:main_results_refined}
\centering
\small
\renewcommand{\arraystretch}{1.05}
\setlength{\tabcolsep}{3pt}
\begin{tabular}{llccc}
\toprule
\textbf{Dataset} & \textbf{Method} & \textbf{2 Agents} & \textbf{3 Agents} & \textbf{4 Agents} \\
\midrule
\multicolumn{5}{l}{\textit{Llama-3.1-8B-Instruct (Accuracy \%)}} \\
\midrule
\multirow{6}{*}{GSM8K}
  & Full Prefill    & 85.97 & 84.69 & 84.69 \\
  & Zero Reuse      & 40.33 & 40.03 & 53.68 \\
  \cmidrule(lr){2-5}
  & CacheBlend      & 76.72 & 77.56 & 74.73 \\
  & EPIC            & 58.73 & 66.46 & 58.42 \\
  & KVComm          & \underline{84.23} & \underline{82.34} & \underline{84.91} \\
  & \cellcolor{cyan!10}RelayCaching
      & \cellcolor{cyan!10}\textbf{84.84}
      & \cellcolor{cyan!10}\textbf{85.50}
      & \cellcolor{cyan!10}\textbf{85.17} \\
\midrule
\multirow{6}{*}{MMLU}
  & Full Prefill    & 71.24 & 73.20 & 66.67 \\
  & Zero Reuse      & 67.97 & 41.18 & 35.29 \\
  \cmidrule(lr){2-5}
  & CacheBlend      & \textbf{69.28} & 64.71 & \underline{66.67} \\
  & EPIC            & \textbf{69.28} & 45.10 & 43.79 \\
  & KVComm          & \underline{68.63} & \underline{69.28} & 66.01 \\
  & \cellcolor{cyan!10}RelayCaching
      & \cellcolor{cyan!10}\textbf{69.28}
      & \cellcolor{cyan!10}\textbf{71.90}
      & \cellcolor{cyan!10}\textbf{67.97} \\
\midrule
\multicolumn{5}{l}{\textit{Qwen2.5-Coder-7B-Instruct (Pass@1 \%)}} \\
\midrule
\multirow{6}{*}{HumanEval}
  & Full Prefill    & 82.61 & 85.71 & 87.58 \\
  & Zero Reuse      & 81.99 & 85.09 & 86.96 \\
  \cmidrule(lr){2-5}
  & CacheBlend      & \underline{82.61} & \underline{84.47} & 85.71 \\
  & EPIC            & 81.99 & 83.23 & \underline{86.34} \\
  & KVComm          & \underline{82.61} & 83.23 & 85.09 \\
  & \cellcolor{cyan!10}RelayCaching
      & \cellcolor{cyan!10}\textbf{82.61}
      & \cellcolor{cyan!10}\textbf{85.09}
      & \cellcolor{cyan!10}\textbf{87.58} \\
\bottomrule
\end{tabular}
\vskip -12pt 
\end{table}

Table~\ref{tab:main_results_refined} and Figure~\ref{fig:reuse_rate_analysis} present comparative results across GSM8K, MMLU, and HumanEval.
RelayCaching consistently matches the Full-Prefill baseline while maintaining reuse rates of over 80\% across most multi-agent settings.
Notably, the layer-range detected on 2WikiMQA generalizes across all tasks without per-task tuning.
Pre-computed caching methods such as EPIC and CacheBlend struggle with the dynamic prefix variations in decode-to-prefill reuse: 
EPIC's static prefix recomputation fails to adapt to content-sensitive deviations, with accuracy dropping to 58.73\% on GSM8K;
CacheBlend's value-norm--based detection does not capture the cosine similarity that governs decode-to-prefill KV deviation patterns, resulting in accuracy below 78\% on GSM8K and MMLU.
KVCOMM achieves competitive accuracy but relies on similarity-based retrieval from anchor pools, causing reuse rates to decline from 88\% to 70\% in GSM8K as context diversity grows.
RelayCaching maintains stable reuse from two to four agents while matching or exceeding KVCOMM's accuracy.
In summary, RelayCaching bridges the accuracy--efficiency gap, achieving Full-Prefill accuracy with over 80\% reuse rate.

\subsection{Efficiency Evaluation}
\label{subsec:efficiency}

We evaluate the system efficiency of RelayCaching on the AIME dataset using Qwen3-0.6B. Our analysis focuses on TTFT acceleration and its scalability across varying numbers of agents and reused sequence lengths.

\textbf{Latency Breakdown Analysis.}
Table~\ref{tab:ttft_breakdown} decomposes the TTFT latency. 
As context accumulates from Agent~2 to Agent~5, full prefill latency increases from 85.2\,ms to 493.9\,ms (5.8$\times$), reflecting the quadratic complexity of prefill computation.
In contrast, RelayCaching's latency grows from 40.6\,ms to only 104.8\,ms (2.6$\times$), achieving a 4.71$\times$ speedup at Agent~5.
%s
The breakdown reveals favorable scaling characteristics.
Index selection grows minimally from 0.5 to 0.8\,ms across agents, validating the efficiency of our token selector.
The full-recompute cost (1.6--7.9\,ms) represents the primary overhead our method yet remains negligible compared to full prefill.
Overall, RelayCaching's latency scales at 2.6$\times$, compared to 5.8$\times$ for Full Prefill and 3.5$\times$ for CacheBlend, demonstrating sub-linear scaling as the number of agents increases.

\begin{figure*}[t]
  \centering
  \captionsetup[subfigure]{skip=1pt, belowskip=-5pt}
  \begin{subfigure}{0.32\textwidth}
    \centering
    \includegraphics[width=\linewidth]{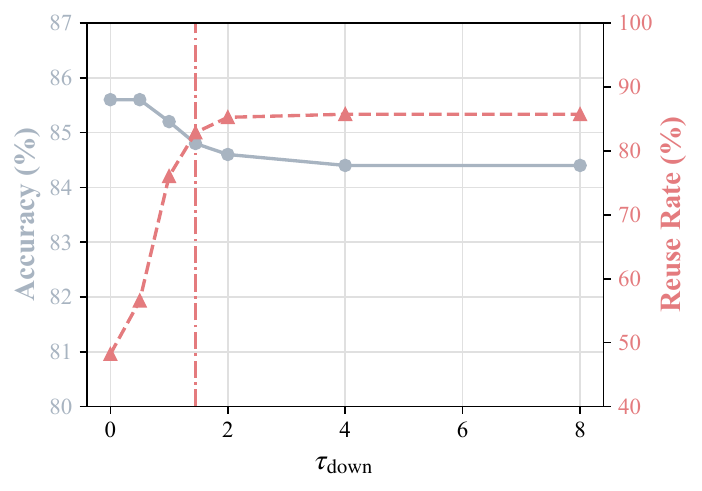}
    \caption{Impact of Influence Score Factor}
    \label{fig:ablation_downstream}
  \end{subfigure}
  \hfill 
  \begin{subfigure}{0.32\textwidth}
    \centering
    \includegraphics[width=\linewidth]{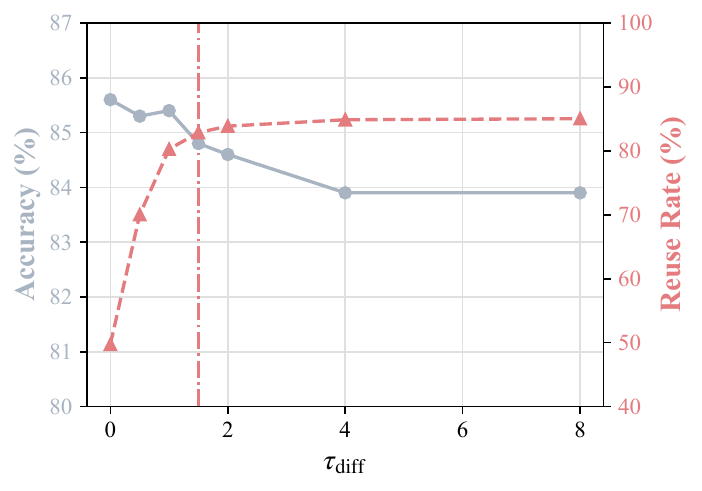}
    \caption{Impact of Deviation Score Factor}
    \label{fig:ablation_difference}
  \end{subfigure}
  \hfill
  \begin{subfigure}{0.32\textwidth}
    \centering
    \includegraphics[width=\linewidth]{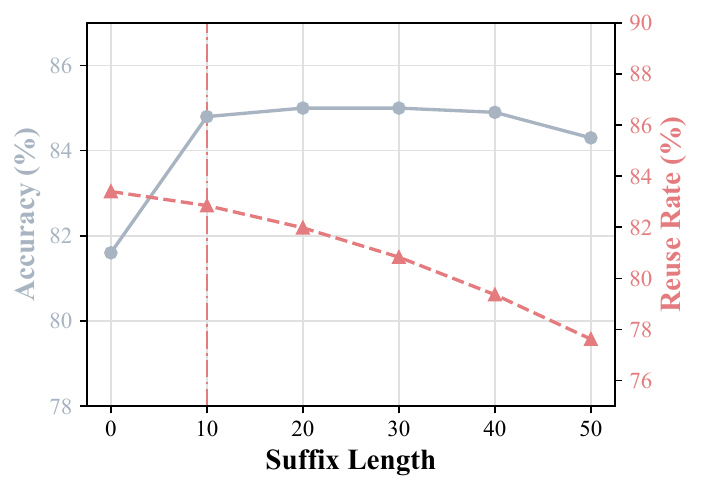}
    \caption{Impact of Suffix Count}
    \label{fig:ablation_suffix}
  \end{subfigure}

\setlength{\belowcaptionskip}{-10pt}

  \caption{\textbf{Ablation on key RelayCaching hyperparameters.} 
  We analyze the trade-off between Accuracy (left axis) and Reuse Rate (right axis) by varying:
  \textit{(a)} the influence score factor $\tau_{\text{down}}$, 
  \textit{(b)} the deviation score factor $\tau_{\text{diff}}$, and 
  \textit{(c)} the suffix length $L_{\text{suf}}$. 
  Red markers denote the default configuration used in the main experiments.}
  \label{fig:hyperparameters}
\end{figure*}

\begin{table}[t]
\centering
\small
\renewcommand{\arraystretch}{1.05}
\caption{\textbf{Per-agent TTFT breakdown and speedup.} Prefix: 512 tokens; Output: 2,048 tokens; Model: Qwen3-0.6B.}
\label{tab:ttft_breakdown}
\begin{tabularx}{\linewidth}{@{}l*{4}{>{\centering\arraybackslash}X}@{}}
\toprule
\textbf{Method} & \textbf{Agent 2} & \textbf{Agent 3} & \textbf{Agent 4} & \textbf{Agent 5} \\
\midrule
Full Prefill & 85.2 & 180.2 & 321.5 & 493.9 \\
\midrule
CacheBlend & 44.9 & 76.7 & 112.5 & 159.3 \\
Speedup & 1.90$\times$ & 2.35$\times$ & 2.86$\times$ & 3.10$\times$ \\
\midrule
RelayCaching & 40.6 & 57.0 & 73.2 & 104.8 \\
\quad Full recompute & 1.6 & 2.8 & 4.3 & 7.9 \\
\quad Index selection & 0.5 & 0.6 & 0.7 & 0.8 \\
\rowcolor{cyan!10}
Speedup & \textbf{2.10$\times$} & \textbf{3.16$\times$} & \textbf{4.39$\times$} & \textbf{4.71$\times$} \\
\bottomrule
\end{tabularx}
\vskip -15pt
\end{table} 

\begin{figure}[t]
  \centering
  \includegraphics[width=0.7\columnwidth]{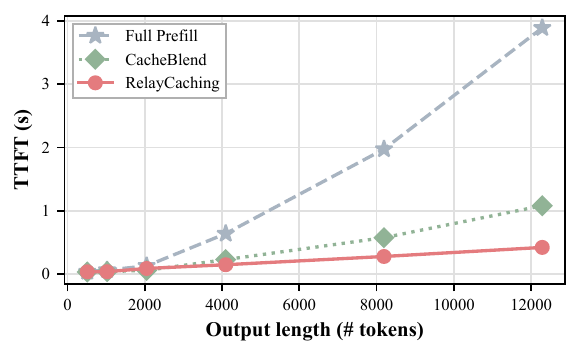}
  \setlength{\abovecaptionskip}{0pt} 
\setlength{\belowcaptionskip}{-10pt}
  \caption{\textbf{Scalability of TTFT with context growth.} 
  We measure the average TTFT of downstream agents as the cumulative output length scales from 512 to 12,288 tokens.}
  \label{fig:ttft_scaling}
\end{figure}

\textbf{Scalability with Context Length.}
To assess performance in long-context regimes, we sweep the cumulative output length from 512 to 12{,}288 tokens among three collaborating agents. 
Figure~\ref{fig:ttft_scaling} reports the average per-agent TTFT.
Standard full prefill exhibits quadratic complexity, with latency escalating to 3.89\,s at 12k tokens.
While CacheBlend mitigates this growth, it still reaches 1.08\,s because it recomputes a fixed proportion of tokens, causing its cost to scale linearly with context length.
In contrast, RelayCaching exhibits sub-linear scaling, recording only 0.42\,s at the maximum length—a 9.2$\times$ speedup over full prefill and 2.5$\times$ over CacheBlend.
This sub-linear scalability stems from the sparsity of token-wise 
deviations (Section~\ref{subsec:obs_token}): as context grows, the number of high-deviation tokens increases more slowly than the total sequence length, keeping the rectification cost sub-linear.

\subsection{Ablation Results}
\label{subsec:ablation}

We conduct a fine-grained ablation on GSM8K under 2 agents setting (Table~\ref{tab:ablation_gsm8k_metrics}).
Restricting rectification to the critical layer range yields 85.90\% accuracy but a modest 46.88\% reuse rate. 
Adding deviation-based selection $\mathbf{I}_{\text{dev}}$ boosts reuse to 90.37\% but degrades accuracy to 79.68\%, indicating that deviation metrics alone overlook generation-critical tokens.
Further incorporating score-based influence selection $\mathbf{I}_{\text{inf-score}}$ improves accuracy to 81.65\% at the cost of lower reuse (85.84\%), confirming that influence-aware criteria capture tokens missed by deviation alone.
The full RelayCaching, integrating all three strategies, achieves 84.84\% accuracy while maintaining 85.35\% reuse.
This shows that multi-dimensional selection is essential for 
preserving generation quality.

\begin{table}[t]
    \centering
    \caption{Ablation study on GSM8K (Llama-3.1-8B-Instruct).}
    \label{tab:ablation_gsm8k_metrics}
    \small
    \renewcommand{\arraystretch}{1.05}
    \begin{tabular}{lcc}
        \toprule
        \textbf{Method} & \textbf{Reuse (\%)} & \textbf{Acc (\%)} \\
        \midrule
        Full Prefill & 0.00 & 85.97 \\
        \midrule
        + Layer Range Profiler & 46.88 & 85.90 \\
        + $\mathbf{I}_{\text{dev}}$ & 90.37 & 79.68 \\
        + $\mathbf{I}_{\text{inf-score}}$ & 85.84 & 81.65 \\
        \rowcolor{cyan!10}
        \textbf{+ $\mathbf{I}_{\text{inf-suffix}}$ (Ours)} & \textbf{85.35} & \textbf{84.84} \\
        \bottomrule
    \end{tabular}
    \vskip -20pt
\end{table}

\subsection{Sensitive Analysis}

We study the impact of the main RelayCaching hyperparameters on the accuracy–reuse trade-off on GSM8K with 2 agents, as shown in Figure~\ref{fig:hyperparameters}. Each panel varies one hyperparameter while keeping the others fixed at their default values.
Panels~(a) and~(b) exhibit consistent patterns: increasing either $\tau_{\text{down}}$ or $\tau_{\text{diff}}$ causes the reuse rate to rise sharply before plateauing, while accuracy varies by 2\%. This indicates that the majority of tokens contribute minimally to deviation correction. Since the steepest trade-off occurs between values of 1 and 2, we select $\tau_{\text{down}} = 1.45$ and $\tau_{\text{diff}} = 1.5$ to maximize reuse with minimal accuracy degradation.
Panel~(c) varies the suffix length $L_{\text{suf}}$ used for recovery. A modest suffix length (e.g., 10) suffices to recover most of the deviation in suffix position, while longer suffixes yield diminishing returns at the cost of reduced reuse.

%% file: contents/6-conclusion.tex
\section{Conclusion}
We introduced RelayCaching, a training-free framework that accelerates multi-agent LLM collaboration by reusing decoding KV caches for subsequent prefilling. 
Our systematic analysis revealed that decoding KV caches remain highly aligned with their full-prefill counterparts despite prefix variation, with value cosine similarity serving as the primary deviation indicator. 
Residual deviations exhibit systematic patterns: a U-shaped layer-wise similarity profile where middle layers show the largest deviations and dominate subsequent generation quality, and token-wise sparsity with strong inter-layer correlation. 
Leveraging these insights, RelayCaching employs a layer-range profiler to confine rectification to a critical layer range, and a token selector combining deviation-based and influence-based criteria to identify a sparse set of tokens requiring rectification. 
Experiments across reasoning, coding, and knowledge benchmarks demonstrate RelayCaching achieves over $80$\% KV cache reuse and up to $4.7\times$ TTFT speedup while preserving accuracy comparable to full prefilling, confirming the effectiveness of decoding-to-prefill KV reuse for efficient multi-agent collaboration.